\let\TeXyear\year
\documentclass{ieeeaccess}
\let\setyear\year
\let\year\TeXyear

\usepackage{lineno,hyperref}

\usepackage{amssymb}
\usepackage{amsmath}
\usepackage{siunitx}
\usepackage{tabularx}
\usepackage{algorithmic}

\usepackage{algorithm2e} 
\usepackage{amsthm}
\usepackage{graphicx}
\usepackage{lscape}
\usepackage{verbatim}
\usepackage{color,soul}
\usepackage{tabularx,ragged2e,booktabs,caption,array,multirow,multicol}
\usepackage{csquotes}
\usepackage{mathtools}
\usepackage{lineno}
\usepackage{amsthm}  
\usepackage{listings}
\usepackage{amssymb}
\usepackage{latexsym}
\usepackage{epsfig}
\usepackage{float}
\usepackage{url}
\usepackage{xspace} 

\usepackage{blindtext}
\usepackage{hyperref}

\usepackage{subcaption}

\usepackage{textcomp}

\usepackage{listings}
\usepackage{lipsum}

\usepackage{cite}
\usepackage{amsmath,amssymb,amsfonts}
\usepackage{algorithmic}
\usepackage{graphicx}
\usepackage{tikz}
\usetikzlibrary{shapes,arrows,calc}
\usepackage{pgfplots}

\usepackage{textcomp}
\def\BibTeX{{\rm B\kern-.05em{\sc i\kern-.025em b}\kern-.08em
    T\kern-.1667em\lower.7ex\hbox{E}\kern-.125emX}}

\definecolor{codegreen}{rgb}{0,0.6,0}
\definecolor{codegray}{rgb}{0.5,0.5,0.5}
\definecolor{codepurple}{rgb}{0.58,0,0.82}
\definecolor{backcolour}{rgb}{0.95,0.95,0.92}

\NewSpotColorSpace{PANTONE}
\AddSpotColor{PANTONE} {PANTONE3015C} {PANTONE\SpotSpace 3015\SpotSpace C} {1 0.3 0 0.2}
\SetPageColorSpace{PANTONE}%

\lstdefinestyle{mystyle}{
    backgroundcolor=\color{backcolour},   
    commentstyle=\color{codegreen},
    keywordstyle=\color{magenta},
    numberstyle=\tiny\color{codegray},
    stringstyle=\color{codepurple},
    basicstyle=\ttfamily\footnotesize,
    breakatwhitespace=false,         
    breaklines=true,                 
    captionpos=b,                    
    keepspaces=true,                 
    numbers=left,                    
    numbersep=5pt,                  
    showspaces=false,                
    showstringspaces=false,
    showtabs=false,                  
    tabsize=2
}

\colorlet{pink}{red!40}
\colorlet{blue}{cyan!60}

 \newcommand{\argmin}{\arg\!\min}

\pgfmathdeclarefunction{gauss}{2}{%
  \pgfmathparse{1/(#2*sqrt(2*pi))*exp(-((x-#1)^2)/(2*#2^2))}%
}

\setyear{2020}
 \pgfplotsset{compat=1.18} 
 
\begin{document}
\doi{https://doi.org/10.48550/arXiv.2304.02595}

\title{Bayesian  neural networks via MCMC: a Python-based tutorial}

\author{\uppercase{Rohitash Chandra}\authorrefmark{1,2}, \IEEEmembership{SM, IEEE}, and Joshua Simmons\authorrefmark{3}} 
\address[1]{Transitional Artificial Intelligence Research Group, School of Mathematics and Statistics, UNSW Sydney,   Australia (e-mail: rohitash.chandra@unsw.edu.au)}
\address[2]{Pingala Institute of Artificial Intelligence, Sydney,   Australia}
\address[3]{ARC Training Centre in Data Analytics for Resources and Environments (DARE), University of Sydney, Australia (e-mail: joshua.simmons@sydney.edu.au}

\tfootnote{This work was supported in part by the  Australian Research Council through Grant IC190100031. We thank Royce Chen from UNSW Sydney for his contribution to the earlier version of this paper. }





\begin{abstract}Bayesian inference provides a methodology for parameter estimation and uncertainty quantification in machine learning and deep learning methods. Variational inference and Markov Chain Monte-Carlo (MCMC) sampling methods  are used to implement Bayesian inference. In the past three decades, MCMC sampling methods have faced some challenges in being adapted to larger models (such as in deep learning) and big data problems. Advanced proposal distributions that incorporate gradients, such as a Langevin proposal distribution, provide a means to address some of the limitations of MCMC sampling for Bayesian neural networks. Furthermore, MCMC methods have typically been constrained to statisticians and currently not well-known among deep learning researchers. We present a tutorial for MCMC methods that covers simple Bayesian linear and logistic models, and Bayesian neural networks. The aim of this tutorial is to bridge the gap between theory and implementation via coding, given a general sparsity of libraries and tutorials to this end. This tutorial provides code in Python with data and instructions that enable their use and extension. We provide results for some benchmark problems showing the strengths and weaknesses of implementing the respective Bayesian models via MCMC. We highlight the challenges in sampling multi-modal posterior distributions for the case of Bayesian neural networks and the need for further improvement of convergence diagnosis methods.
\end{abstract}


\titlepgskip=-15pt

\maketitle 

\subsection{Introduction}
\label{sec:introduction}

\PARstart{B}{ayesian}  inference provides a probabilistic approach for parameter estimation in a wide range of models used across the fields of machine learning, econometrics, environmental and Earth sciences  \cite{chamberlain2003nonparametric, geweke1989bayesian,box2011bayesian,chandra2019PT-Bayeslands,JPall_BayesReef2020}. The term 'probabilistic' refers to the representation of unknown parameters as probability distributions rather than using fixed point estimates as in conventional machine learning models where gradient-based optimisation methods are prominent \cite{kingma2014adam}. A probabilistic representation of unknown parameters requires a different approach to optimisation, which is known as \textit{sampling} from a  computational statistics point-of-view \cite{martino2014metropolis}.

Markov Chain Monte-Carlo (MCMC) sampling methods have been prominent for inference (estimation) of model parameters via the posterior probability distribution. In other words, Bayesian methods attempt to quantify the uncertainty in model parameters by marginalising over the predictive posterior distribution. Hence, in the case of neural networks, MCMC methods can be used to implement Bayesian neural networks that represent weights and biases as probability distributions \cite{mackay1995probable,neal2012bayesian,specht1990probabilistic,richard1991neural,wan1990neural}. 
Probabilistic machine learning provides  natural  way of providing uncertainty quantification in predictions  \cite{ghahramani2015probabilistic}, since the uncertainties can be obtained by probabilistic representation of parameters. This inference procedure can be seen as a form of learning (optimisation) applied to the model parameters \cite{neal2012bayesian}. In this tutorial, we employ linear models and simple neural networks to demonstrate the use of MCMC sampling methods. The probabilistic representation of weights and biases in the respective models allows uncertainty quantification on model predictions.

We note that MCMC refers to a family of algorithms for implementing Bayesian inference for parameter and uncertainty estimation in models. Bayesian inference applications include statistical, graphical, and machine learning models. The differences in the model complexity from different domains have led to the existence of a wide range of MCMC sampling algorithms. Some of the prominent ones are Metropolis-Hastings algorithm \cite{chib1995understanding,robert1999metropolis,hitchcock2003history}, Gibbs sampling \cite{
casella1992explaining,carter1994gibbs,roberts1994simple}, Langevin MCMC \cite{rossky1978brownian,roberts1996exponential,roberts1998optimal},  rejection sampling \cite{flury1990acceptance,gilks1992adaptive}, importance sampling \cite{tokdar2010importance,llorente2021deep}, sequential MCMC \cite{brockwell2010sequentially},  adaptive MCMC \cite{andrieu2008tutorial}, parallel tempering (tempered) MCMC \cite{swendsen1986replica,hukushima1996exchange,earl2005parallel,sambridge2014parallel}, reversible-jump MCMC \cite{
fan2011reversible,brooks2003efficient}, specialised MCMC methods for discrete time series models\cite{tarantola2004mcmc,liang2005efficient,zanella2020informed}, constrained parameter and model settings \cite{browne2006mcmc,gallant2022constrained}, and likelihood free MCMC \cite{sisson2011likelihood}. MCMC sampling methods have also been used for data augmentation \cite{van2001art,duan2018scaling}, model fusion \cite{zobitz2011primer}, model selection \cite{andrieu2001model,andrieu1999joint}, and interpolation \cite{mugglin1999bayesian}. Apart from this, we note that MCMC methods have been prominent in a wide range of applications that include geophysical inversions \cite{sambridge1999geophysical,sambridge2002monte,scalzo2019efficiency}, geoscientific models \cite{JPall_BayesReef2020,chandra2019bayeslands,olierook2021bayesian}, environmental and hydrological modelling \cite{marshall2004comparative,vrugt2008treatment},  bio-systems modelling  \cite{valderrama2019mcmc,nishiyama2018interaction,rannala2002identifiability}, and quantitative genetics \cite{sorensen2002likelihood,drummond2012bayesian}.

In the case of Bayesian neural networks, the large number of model parameters that  emerge from large neural network architectures and deep learning models pose challenges for MCMC sampling methods. Hence, progress in the application of Bayesian approaches to big data and deep neural networks has been slow. Research in this space has included a number of methods that have been fused with MCMC such as gradient-based methods \cite{girolami2011riemann,roberts1998optimal,  neal2011mcmc,welling2011bayesian,Chandra2019NC}, and evolutionary (meta-heuristic) algorithms which include differential evolution, genetic algorithms, and particle swarm optimisation  \cite{drugan2004evolutionary,ter2006markov,ter2008differential, kapoor2022bayesian}. 

This use of gradients in MCMC was initially known as Metropolis-adjusted Langevin dynamics \cite{roberts1998optimal} and has shown promising performance for linear models \cite{welling2011bayesian} and has also been extended to Bayesian neural networks \cite{Chandra2019NC}.  
Hamiltonian Monte Carlo (HMC) sampling also employ gradient-based proposal distributions \cite{neal2011mcmc} and has been effectively applied to Bayesian neural networks \cite{li2019neural}. In similar way, Langevin dynamics can be used to incorporate gradient-based stepping with Gaussian noise into the proposal distribution \cite{welling2011bayesian}. HMC avoids random walk behaviour using an auxiliary momentum vector and implementing Hamiltonian dynamics where the momentum samples are discarded  later. The samples are hence less correlated and tend converge to the target distribution more rapidly. Another direction has been the use of better exploration features in MCMC sampling such as parallel tempering MCMC with Langevin proposal distribution and parallel computing \cite{Chandra2019NC}. These have the ability to provide a  competitive alternative to  stochastic gradient-descent \cite{bottou2010large} and Adam optimizers \cite{kingma2014adam} with the addition of uncertainty quantification in predictions. These methods have also been applied to Bayesian deep learning models such as Bayesian autoencoders \cite{chandra2021revisiting} and Bayesian graph convolutional neural networks (CNNs) \cite{chandra2021bayesian} which require millions of trainable parameters to be represent as posterior distributions. Recently, Kapoor et al. \cite{kapoor2022bayesian} combined tempered MCMC with particle swarm optimisation-based proposal distribution in a parallelized environment that showed more effective sampling when compared with the conventional approach. However, we note that large deep learning models can feature hundreds of millions to billions of parameters, which brings further challenges to sampling strategies and hence the road is less travelled.

 Variational inference  provides an alternative approach to MCMC methods  to  approximate Bayesian posterior distribution \cite{blei2017variational,Graves2011Variational}. \textit{Bayes by backpropagation} is a VI method that showed competitive results when compared to stochastic gradient descent and \textit{dropout} methods used as approximate Bayesian methods \cite{blundell2015weight}. Dropout is a regularisation technique that  involves randomly dropping selected weights in forward-pass operation of backpropagation. This improves the generalization performance of neural networks and has been widely adopted \cite{srivastava2014dropout}. Gal and Ghahramani \cite{gal2016dropout} presented an approximate  Bayesian methodology  based on dropout-based regularisation which has been used for other deep learning models such as  CNNs \cite{shridhar2019comprehensive}. Later, Gal and Ghahramani \cite{gal2016theoretically} presented variational inference-based dropout technique for recurrent neural networks (RNNs); particularly, long-short term memory (LSTM) and gated recurrent unit (GRU) models for  language modelling and sentiment analysis tasks.
 
We argue that the use of dropouts for Bayesian inference \cite{shridhar2019comprehensive} cannot be seen as an alternative to MCMC sampling which samples directly from the posterior distribution. In the case of dropouts for Bayesian inference, we do not know the priors nor know much about the posterior distribution and there is little theoretical rigour, only computational efficacy or capturing noise and uncertainty during model training. Furthermore, in the Bayesian methodology, a probabilistic representation using priors is needed which is questionable in the dropout methodology for Bayesian computation. Given that variational inference methods are seen as approximate Bayesian methods, we need to invest more effort in directly sampling from the posterior distribution  for Bayesian deep learning models. This can only be possible if both communities (i.e., statistics and machine learning) are aware about the strengths and weaknesses of MCMC methods for sampling Bayesian neural networks that span hundreds to thousands of parameters, and go orders of magnitude higher when looking at Bayesian deep learning models. The progress of MCMC for deep learning has been slow, due to lack of implementation details, libraries and tutorials that provide that balance of theory and implementation.

In this paper, we present a Python-based MCMC sampling tutorial  for simple Bayesian linear  models and Bayesian neural networks. We provide code in Python with data and instructions that enable their use and extension. We provide detailed instructions for   sample code in a related Github repository which is easy to clone and run. Our code implementation is simple and relies on basic Python libraries such as \textit{
numpy} as the goal of this tutorial is to serve as a go-to document for beginners who have basic knowledge of machine learning models, and need to get hands-on experience with MCMC sampling. Hence, this is a code-based computational tutorial with a theoretical background. We provide results for some benchmark problems showing the strengths and weaknesses of implementing the respective Bayesian models. Finally, we highlight the challenges in sampling multi-modal posterior distributions in the case of Bayesian neural networks and shed light on the use of convergence diagnostics. 
 
The rest of the paper is organised as follows. In Section 2, we present a background and literature review of related methods. Section 3 presents the proposed methodology that includes MCMC sampling code implementation in Python for respective models, followed by experiments and results in Section 4. Section 5 provides a discussion and Section 6 concludes the paper with directions for future work. 

\subsection{Background} 

\subsubsection{Bayesian inference}
\label{sec:background-binf}

\begin{figure}[!htb]
    \centering
    \begin{tikzpicture}[
        box/.style={rectangle, draw, text width=4em, text centered, minimum height=2.5em, font=\footnotesize},
        widebox/.style={rectangle, draw, text width=10em, text centered, minimum height=2.5em, font=\footnotesize},
        line/.style={draw, -latex'}
    ]
    
    \node [box, align=center] (data) {Data\\$d$};
    \node [box, right of=data, node distance=2cm, align=center] (prior) {Prior\\$P(\Theta)$}; 
    \node [box, left of=data, node distance=2cm, align=center] (likelihood) {Likelihood\\$P(d|\Theta)$};
    \node [widebox, below of=data, node distance=1.5cm, align=center] (bayes) {Bayes' Theorem\\$P(\Theta|d) \propto P(d|\Theta)P(\Theta)$};
    \node [box, below of=bayes, node distance=1.5cm] (posterior) {Posterior\\$P(\Theta|d)$};
    
    \path [line] (likelihood.south) -- ++(0,-0.25) -| (bayes.north);
    \path [line] (data.south) -- ++(0,-0.25) -| (bayes.north);
    \path [line] (prior.south) -- ++(0,-0.25) -| (bayes.north);
    \path [line] (bayes) -- (posterior);
    
    \end{tikzpicture}
    \caption{ We show the relationship of the likelihood with data and prior distribution for sampling the posterior distribution.}
    \label{fig:bayes}
\end{figure}
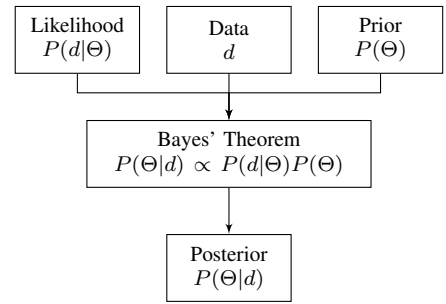

We recall that Bayesian methods account for the uncertainty in prediction and decision-making via the posterior distribution \cite{bernardo2009bayesian}. Note that the posterior is the conditional probability determined after taking into account the prior distribution and the relevant evidence or data via sampling methods. Thomas Bayes  (1702 – 1761) presented and proved a special case of the Bayes' theorem \cite{barnard1958studies,dale2012history} which is the foundation of Bayesian inference. However, it was Pierre-Simon Laplace (1749 – 1827) who introduced a general version of the theorem and used it to approach problems \cite{laplace1986memoir}. Figure \ref{fig:bayes} gives an overview of the Bayesian inference framework that uses data with a prior and likelihood to construct to sample from the posterior distribution. This is the building block of the rest of the lessons that will feature Bayesian logistic regression and Bayesian neural networks.

Bayesian inference estimates unknown parameters using prior information or belief about the variable. Prior information is captured in the form of a distribution. A simple example of a prior belief is a distribution that has a positive real-valued number in some range. This essentially would imply a belief that our result or posterior distribution would likely be a distribution of positive numbers in some range which would be similar to the prior but not the same. If the posterior and prior both follow the same type of distribution, this is known as a \textit{conjugate prior} \cite{johnson2010methods}. If the prior provides useful information about the variable, rather than very loose constraints, it is known as an \textit{informative prior}. The prior distribution is based on expert knowledge (opinion) and is also dependent on the domain for different types of models \cite{banner2020use,van2021bayesian}. 


The need for efficient sampling methods to implement Bayesian inference has been a significant focus of research in computational statistics. This is especially true in the case of multimodal and irregular posterior distributions \cite{li2015adaptive,hu2012bayesian,sambridge2014parallel} which tend to dominate  Bayesian neural networks \cite{richard1991neural,rojas1996short}. MCMC sampling methods are used to update the probability for a hypothesis (proposal $\Theta$) as more information becomes available. The hypothesis is given by a prior probability distribution that expresses one's belief about a quantity (or free parameter in a model) before some data ($\bf{d}$) are observed. MCMC sampling methods construct the posterior distribution ($P(\Theta | \bf{d})$) iteratively, using a proposal distribution, prior distribution $P(\Theta)$ and a likelihood function ($P(\bf{d}| \Theta)$), as expressed below

\begin{equation}
 P(\Theta | \bf{d}) = \frac{ P(\bf{d} | \Theta ) \times P(\Theta)}{P(d)}.
\end{equation}

$P(\bf{d})$  is the marginal distribution of the data and is often seen as a normalising constant and ignored. Hence, ignoring it, we can also express the above by

\begin{equation}
 P(\Theta | \bf{d}) \propto P(\bf{d} | \Theta ) \times P(\bf{\Theta}).
\end{equation}

The likelihood function is a function of the parameters of a given model provided specific observed data \cite{akaike1998likelihood}. The likelihood function can be seen as a measure of fit to the data, given the proposals that are drawn from the proposal distribution. Hence, from an optimisation perspective, the likelihood function can be seen as a fitness or error function. The posterior distribution is constructed after taking into account the relevant evidence (data) and prior distribution, with the likelihood that considers the proposal and the model. MCMC methods essentially implement Bayesian inference via a numerical approach that marginalizes or integrates over the posterior distribution \cite{mackay1996hyperparameters}. Note that \textit{probability} and \textit{likelihood} are not the same in the field of statistics, while in everyday language they are used as if they are the same.  The term "probability" refers to the possibility of something happening, in relation to a given distribution of data. The likelihood refers to the likelihood function that provides a measure of fitness in relation to a distribution. The likelihood function indicates which parameter (data) values are more likely than others in relation to a distribution. Further and detailed explanations regarding Bayesian inference and MCMC sampling have been  given in the literature \cite{gelman-bayesian-2004,mcelreath-statistical-2020}.

\subsubsection{Probability distributions}
\label{sec:prob_dists}

\paragraph{Gaussian (Normal) Distribution}
A normal probability density or distribution, also known as the Gaussian distribution, is described by two parameters, mean ($\mu$) around which the distribution is centered and the standard deviation ($\sigma$) which describes the spread (sometimes described instead by the variance, $\sigma^2$). Using these two parameters, we can fit a probability (normal) distribution to data from some source. In a similar way, given a probability distribution, we can generate data and this process is known as sampling from the distribution. In sampling the distribution, we simply present random data points (uniform) to the distribution and get data that are, in a way, transformed by the distribution. These parameters determine the shape of the probability distribution, e.g., if it is peaked or spread. Note that the normal distribution is symmetrical and caters for negative and positive numbers of real data. 

Equation \ref{eqn:gauss_dist} presents the Gaussian distribution probability density function (PDF) for parameters $\mu$ and $\sigma$.

\begin{equation}
    f(x)=\frac{1}{\sqrt{2\pi\sigma^2}}exp\left({-\frac{1}{2}\left(\frac{x-\mu}{\sigma}\right)^2}\right)
\label{eqn:gauss_dist}
\end{equation} 

\begin{figure}[!htb]
    \centering
    \resizebox{\linewidth}{!}{\input{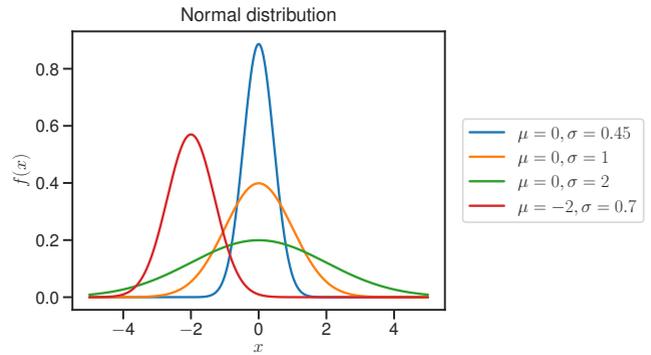}}
    \caption{Normal distributions with different parameters, i.e., mean and the standard deviation.}
    \label{fig:normal_distribution_different_parameters}
\end{figure}

We will sample from this distribution in Python via the NumPy library \cite{harris2020array} which covers the various distributions discussed in this tutorial. The SciPy library \cite{2020SciPy-NMeth} is used to get a representation of  the probability distribution function (PDF). The associated Github repository\footnote{\url{https://github.com/sydney-machine-learning/Bayesianneuralnetworks-MCMC-tutorial/blob/main/01-Distributions.ipynb}} contains the code to generate Figures \ref{fig:normal_distribution_different_parameters} to \ref{fig:inverse_gamma_distribution} using the Seaborn and Matplotlib Python libraries. Listing \ref{lst:gaussian-dist} shows an example of drawing samples from a Gaussian distribution and obtaining the PDF values across the domain 0 to 8. The SciPy and NumPy libraries cover all distributions mentioned below.

\begin{lstlisting}[language=Python, label={lst:gaussian-dist}, caption=Random number generation for a Gaussian distribution]
import numpy as np
from numpy import random
from scipy import stats

# 10 random draws from a standard Gaussian distribution - N(0,1)
samples = random.randn(10)
# Gaussian distribution PDF with mean of 4 and standard deviation of 0.5
x = np.linspace(0, 8, 100)
pdf = stats.norm.pdf(x, loc=4, scale=0.5)
\end{lstlisting}

We note that the mean and standard deviation are purely based on the data that will change depending on the dataset. Let us visualise what happens to the distribution when the mean and standard deviation change, as shown in Figure \ref{fig:normal_distribution_different_parameters}.



\paragraph{Multivariate Normal distribution}

The multivariate normal distribution or joint normal distribution generalises univariate normal distribution to more variables or higher dimensions, as shown in the PDF in Equation \ref{eqn:mvn_dist}.

\begin{align}
    f(x_1,\ldots,x_M) =&\notag\\ \frac{1}{\sqrt{(2\pi)^M|\mathbf\Sigma|}} &\exp\left(-\frac{1}{2} ({\mathbf x}-{\mathbf\mu})^\mathrm{T}{\mathbf\Sigma}^{-1}({\mathbf x}-{\mathbf\mu})\right)
\label{eqn:mvn_dist}
\end{align}

where $\mathbf{x}$ is a real $M$-dimensional column vector and $|\Sigma|$ is the determinant of the  \textit{symmetric covariance matrix}, which is \textit{positive definite}.
 
\paragraph{Gamma distribution}

A gamma distribution is defined by the parameters shape ($\alpha$) and rate ($\beta$), as shown below.

\begin{equation}
    f(x;\alpha,\beta) = \frac{ \beta^\alpha x^{\alpha-1} e^{-\beta x}}{\Gamma(\alpha)}
\label{eqn:gamma}
\end{equation}

for  $x > 0 \quad  \alpha, \beta > 0$; where $\Gamma(n) = (n-1)!$. Figure  \ref{fig:gamma_distribution} presents the Gamma distribution for various parameter combinations, with the corresponding code in the accompanying Github repository.  

\begin{figure}[htbp!]
    \centering
    \resizebox{\linewidth}{!}{\input{figures/gamma_distribution.pgf}}
    \caption{Gamma distributions with different shape and rate parameters ($\alpha$ and $\beta$).}
    \label{fig:gamma_distribution}
\end{figure}

\begin{figure}[htbp!]
    \centering
    \resizebox{\linewidth}{!}{\input{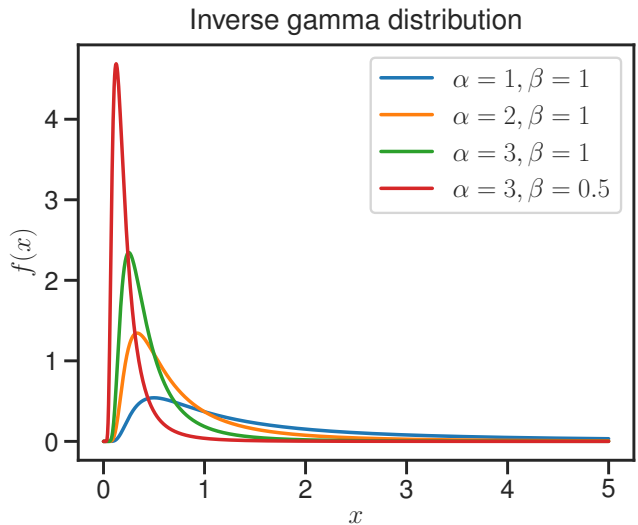}}
    \caption{Inverse gamma distributions with different shape and rate parameters  ($\alpha$ and $\beta$).}
    \label{fig:inverse_gamma_distribution}
\end{figure}

The corresponding inverse-Gamma (IG) distribution takes the same parameters with examples given in Figure \ref{fig:inverse_gamma_distribution} and is more appropriate for real positive numbers.

\paragraph{Binomial distribution}

So far, we have only addressed real numbers with respective probability distributions; however, we also need to consider discrete numbers. The Bernoulli distribution is a discrete probability distribution typically used for modelling binary classification problems. We begin with an example where a variable $x$ takes the value 1 with probability $p$ and the value 0 with probability $q=1-p$. We give the probability mass function for this distribution over the possible outcomes ($x$) in Equation \ref{eqn:binomial}.

\begin{equation}
   f(x;p) = p^x (1-p)^{1-x} 
\label{eqn:binomial}
\end{equation}

for  $x\in{0,1} $. The probability of getting exactly $k$ successes ($x=1$) in $n$ independent Bernoulli trials ($f(k,n,p)$) is given as $\Pr(k;n,p)$ in Equation  \ref{eqn:ber}.

 \begin{equation}
 \Pr(k;n,p) = \binom{n}{k}p^k(1-p)^{n-k} 
 \label{eqn:ber}
\end{equation}
for $k = 0, 1, 2, ..., n$, where $
\binom{n}{k} =\frac{n!}{k!(n-k)!}$.  

\paragraph{Multinomial distribution}

Previously, we catered for the case of two outcomes; however, we can consider the case of more than two outcomes. Suppose a single trial can result in $k \ (k \geq 2)$ possible outcomes numbered $1, 2, \ldots, k$ and let $p_i = \mathbb{P}(\text{a single trial results in outcome i})$ $(\sum_{i=1}^k p_i = 1)$. In the case of $n$ independent trials, let $X_i$ denote the number of trials resulting in outcome $i$ (then $\sum_{i=1}^k X_i = n$). Then, we can state that the distribution of $(X_1, X_2, \ldots, X_k) \sim \text{Multinomial}(n; p_1, p_2, \ldots, p_k)$, and it holds 

\begin{align}
&\mathbb{P}(X_1 = x_1, X_2 = x_2, \ldots, X_k = x_k) =\notag\\ &\frac{n!}{x_1! x_2! \ldots x_k!} p_1^{x_1}p_2^{x_2}\ldots p_k^{x_k}, \  0 < p_i < 1, \ \sum_{i=1}^k p_i = 1 .
\label{eqn:multinominal}
\end{align}

\subsection{MCMC}

We begin by noting that a Markov process is uniquely defined by its transition probabilities  $P(x'| x)$, which defines the probability of transitioning from any given state $x$ to another given state  $x'$. The Markov process has a unique stationary distribution $\pi(x)$ given the following two conditions are met.
 
\begin{enumerate}
    \item  There must exist a stationary distribution $\pi$ which solves the detailed balance equations, and therefore requires that each transition $x \to x'$  is reversible. This implies that for every pair of states $x,x'$, the probability of being in state $x$ and moving to state $x'$, must be equal to the probability of being in state $x'$  and moving to state $x$; hence, $\pi (x)P(x'\mid x)=\pi (x')P(x\mid x')$.  
 
    \item  The stationary distribution must be unique, which is guaranteed by ergodicity of the Markov process \cite{roberts2004general,roberts2006extremal,andrieu2006ergodicity,grazzini2012analysis}. Ergodicity is guaranteed when every state is aperiodic (i.e., the system does not return to the same state at fixed intervals) and positive recurrent (i.e., the expected number of steps for returning to the same state is finite). An ergodic system is one that mixes well, in other words, you get the same result whether you average its values over time or space.             
\end{enumerate}

Given that $\pi(x)$ is chosen to be $P(x)$, the condition of detailed balance becomes $P(x' \mid x) P(x) = P(x \mid x') P(x')$
which is re-written as shown in Equation \ref{eqn:cond}.

 \begin{equation}
\frac{P(x' \mid x)}{P(x \mid x')} = \frac{P(x')}{P(x)}
 \label{eqn:cond}
\end{equation}

 Algorithm \ref{alg:simple_mcmc} presents a basic MCMC sampler with random-walk proposal distribution that runs until a maximum number of samples ($N_{max}$) has been reached for training data, $\mathbf{d}$.
 
\begin{algorithm}

\KwData{Training data, $\mathbf{d}$}
\KwResult{$N_{max}$ samples from the posterior distribution}
 - Initialise $x_{0}$ \;
\For{$i=1$ until $N_{max}$}{
    1. Propose a value $x'|x_i \sim q(x_i)$, where $q(.)$ is the proposal distribution\;
    2. Given $x'$, execute the model $f(x',\mathbf{d})$ to compute the predictions (output $y$) and the likelihood\;
    3. Calculate the acceptance probability 
    $\alpha  =  \min\left(1, \frac{P(x')}{P(x_i)} \frac{q(x_i \mid x')}{q(x' \mid x_i)}\right)$\\
    4. Generate a random value from a uniform distribution $u \sim U(0,1)$\;
    5. Accept or reject proposed value $x'$\;
    \eIf{$u < \alpha$}{
        accept the sample, $x_{i} = x'$  
    }{ 
        reject current  and retain previous sample, $x_{i} = x_{i-1}$  
    }
}

\caption{A basic MCMC sampler leveraging the Metropolis-Hastings algorithm}
\label{alg:simple_mcmc}
\end{algorithm}

Algorithm \ref{alg:simple_mcmc} proceeds by proposing new values of the parameter $x$ (Step 1) from the selected proposal distribution $q(.)$; in this case, a uniform distribution between 0 and 1. Conditional on these proposed values,  the model $f(x', \mathbf{d})$ computes or predicts an output  using proposal x' and data $\mathbf{d}$ (Step 2). We compute the likelihood using the prediction and employ a Metropolis-Hasting criterion (Step 3) to determine whether to accept or reject the proposal (Step 5). We compare  the acceptance ratio $\alpha$  with $u \sim U(0,1)$,  this enforces that the proposal is accepted with probability $\alpha$. If the proposal is accepted, the chain moves to this proposed value. If rejected, the chain stays at the current value. The process is repeated until the convergence criterion is met, which  is the maximum number of samples ($N_{max}$). 

\subsubsection{Priors}

The prior distribution is generally based on belief, expert opinion or other information without viewing the data \cite{neal2012bayesian,cheng1994neural}. Information to construct the prior can be based on past experiments or the posterior distribution of the model for related datasets. There are no hard rules for how much information should be encoded in the prior distribution; hence,  we can take multiple approaches.

An \textit{informative prior} gives specific and definite information about a variable. If we consider the prior distribution for the temperature tomorrow evening, it would be reasonable to use a  normal distribution with an expected value (as mean) of today's evenings temperature with a standard deviation of the temperature each evening for the entire season. A \textit{weakly informative prior} expresses partial information about a variable. In the case of the prior distribution of evening temperature, a weakly informative prior would consider day time temperature of the day (as mean) with a standard deviation of day time temperature for the whole year. An \textit{uninformative prior} or \textit{diffuse prior} expresses vague information about a variable, such as the variable is positive or has some limit range. 

A number of studies have been done regarding priors for linear models \cite{smith1973general,bedrick1996new} and Bayesian neural networks and deep learning models  \cite{fortuin2022priors}. Hobbs et al. \cite{hobbs2012commensurate} presented a study for  Bayesian priors in generalised linear models for clinical trials. We note that incorporation of prior knowledge in deep learning models \cite{de2019deep}, is different from selecting or defining priors in Bayesian deep learning models. Due to the similarity of terms, we caution the readers that these can be often confused and mixed up.

In the case of Bayesian neural networks, the prior distribution can be based on the distribution of the weights and biases from similar neural network models. This can be seen as an example of expert knowledge and implemented in previous studies \cite{chandra_bayesian_2021,chandra2021revisiting}. Another example of expert knowledge is the concept of  \textit{weight decay}  \cite{krogh1992simple} regularisation (L2 or Ridge regression \cite{mcdonald2009ridge})   which restricts large weights and can be incorporated when defining the  prior distribution (priors) \cite{mackay1995probable,mackay1995probable,neal2012bayesian}.


\subsubsection{ MCMC sampler in Python }
\label{sec:basic_mcmc}

We begin with a deliberately simple example where we sample one parameter from a binomial distribution to demonstrate a simple MCMC implementation in Python. Looking at a simple binomial (e.g., coin flipping) likelihood (we will explore the likelihood later),  given the data of $k$ successes in $n$ trials, we calculate the posterior probability of the parameter $p$ that defines the chance of success for any given trial. MCMC sampling requires a prior distribution and a likelihood function to evaluate  a set of parameters (proposed) for the given data and model. In other words, the likelihood is the measure of the quality of proposals obtained from a proposal distribution. 

Listing \ref{lst:simple-mcmc} presents an implementation \footnote{\href{https://github.com/sydney-machine-learning/Bayesianneuralnetworks-MCMC-tutorial/blob/main/02-Basic-MCMC.ipynb}{https://github.com/sydney-machine-learning/Bayesianneuralnetworks-MCMC-tutorial/blob/main/02-Basic-MCMC.ipynb}} of this simple MCMC sampling exercise in Python of Algorithm 1.

In this example, we adopt a uniform distribution as an uninformative prior, only constraining the $p$ to be between the values of 0 and 1 ($p \in [0,1]$).
\begin{figure*}[t]
\begin{lstlisting}[language=Python, label={lst:simple-mcmc}, caption=Python implementation of Algorithm 1]
# First define our likelihood function which will be dependent on provided `data`
# in this case we will choose k = 50, n = 100
def likelihood(query_prob):
    '''
    Given the data of k successes in n trials, return a likelihood function which 
    evaluates the probability that a single success (p) is query_prob for any given 
    query_prob (between 0 and 1).
    '''
    k = 50
    n = 100
    return stats.binom.pmf(k, n, query_prob)

## MCMC Settings and Setup
n_samples = 10000 # number of samples to draw from the posterior
burn_in = 2500 # number of samples to discard before recording draws from the posterior

x = random.uniform(0, 1) # initialise a value of x0

# create an array of NaNs to fill with our samples
p_posterior = np.full(n_samples, np.nan) 

print('Generating {} MCMC samples from the posterior:'.format(n_samples))

# now we can start the MCMC sampling loop
for ii in np.arange(n_samples):
    # Sample a value uniformly from 0 to 1 as a proposal
    x_new = random.uniform(0, 1)

    # Calculate the Metrpolis-Hastings acceptance probability based on the prior 
    # (can be ignored in this case) and likelihood
    prior_ratio = 1 # for this simple example as discussed above
    likelihood_ratio = likelihood_function(x_new) / likelihood_function(x)
    alpha = np.min([1, likelihood_ratio * prior_ratio])

    # Here we use a random draw from a uniform distribution between 0 and 1 as a 
    # method of accepting the new proposal with a probability of alpha
    # (i.e., accept if u < alpha)
    u = random.uniform(0, 1)
    if u < alpha:
        x = x_new # then update the current sample to the propoal for the next iteration

    # Store the current sample
    p_posterior[ii] = x
\end{lstlisting}
\end{figure*}
 
In MCMC sampling, a certain portion of the initial samples are discarded that are known as the burn-in or warmup period which could be seen as an optimisation stage. The burn-in period depends on the sampling problem (complexity of the model). In this simple case,   we will use 25 \% burn-in  and in the case of  neural network models, 50 \% burn-in will likely be required.  Essentially, during burn-in, we are discarding material that is not part of the posterior distribution, since the posterior should feature good proposals which we get once the sampler goes towards convergence. 

Typically, histograms of the posterior distribution and the trace plot are used to visualise the MCMC sampling performance. The histogram of the posterior distribution allows us to examine the mean and variance visually, while the trace plot shows the value of samples at each iteration, allowing us to examine the behaviour and convergence of the MCMC.

\begin{figure}[!htb]
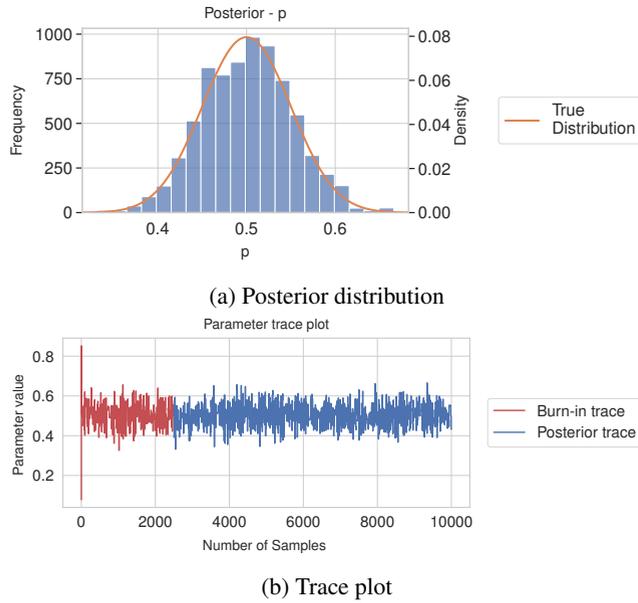

    \centering
    \begin{subfigure}{\linewidth}
        \resizebox{\linewidth}{!}{\input{figures/02-Basic-MCMC_posterior.pgf}}
        \caption{Posterior distribution}
    \end{subfigure}
    \begin{subfigure}{\linewidth}
        \resizebox{\linewidth}{!}{\input{figures/02-Basic-MCMC_trace.pgf}}
    \caption{Trace plot}
    \end{subfigure}
    \caption{Posterior and trace plot for the basic MCMC sampler given in Listing \ref{lst:simple-mcmc}.}
    \label{fig:simplehisto}
\end{figure}

Although it is necessary to exclude the burn-in samples in the posterior distribution, it can be helpful to include them in the trace plots so that we can examine where the model started and how well it converged. We provide a visualisation of the results that features a normal distribution obtained by a simple MCMC sampling from executing code Listing \ref{lst:simple-mcmc}.   The histogram of the posterior shows a normally distributed shape (Figure \ref{fig:simplehisto} - Panel a), and the trace plot  (Figure \ref{fig:simplehisto} - Panel b) shows that the samples are distributed around the convergence value, as well as the burn-in samples which are in red. We also note that the value of the posterior is usually taken as the mean of the distribution, and in this case, the mean value is $0.502$. The median can also be taken as a measure which would be more useful in irregular distributions. 

\subsection{Bayesian linear models via MCMC}
\label{sec:lm_mcmc}

We provide details of implementing \textit{Bayesian linear regression} that uses MCMC sampling with  random-walk proposal distribution.   \textcolor{black}{We wish to model a dataset  consisting of  input (features or covariates)  $\mathbf x=(x_1, \ldots,   x_S)'$  and corresponding outputs $\mathbf y=(y_1,\ldots, y_S)'$  for $S$ instances in data. } \textcolor{black}{This approach models the response observations as being composed of a regression component (the linear regression denoted by $f(\mathbf x, \mathbf \theta)$) and a noise term (Gaussian distribution with a mean of zero ($\mu=0$) and variance $\tau^2$ (see Equation \ref{eqn:base-blr}). In the Bayesian linear regression, we treat the parameters  ($\theta$ and $\tau^2$) as random variables to be estimated (sampled) based on the data and likelihood. So}

\begin{equation}
    \label{eqn:base-blr}
    \bar{\mathbf y}=f(\mathbf x, \mathbf \theta)+e \qquad e\sim \mathcal{N}(0,\tau^2)
\end{equation}
or equivalently, 
\begin{equation}
    \label{eqn:likey}
    p(\mathbf y|\mathbf x,\theta,\tau^2) \sim \mathcal{N}\left(f(\mathbf x,\mathbf \theta),\tau^2\right).
\end{equation}
 

Equation \ref{eqn:linear_model} expresses the general case of a linear model using a vector of input data  $\mathbf x$ to obtain prediction $\mathbf y$.

\begin{equation}
    f(\mathbf x,\theta) = \mathbf \theta \mathbf{x}^T
 \label{eqn:linear_model}
\end{equation}

\textcolor{black}{In the case of Bayesian linear regression, $\mathbf \theta$  is a set of distributions (typically Gaussian) rather than a fixed point estimate in conventional linear models. Therefore, we estimate the parameters ($\mathbf \theta$ and $\tau^2$) using MCMC sampling to obtain their posterior distributions.} Note that from an optimisation perspective, the case of sampling can be seen as a form of optimisation, e.g., using gradient-based methods  for learning the parameters of  linear models and neural networks in the machine learning and neural networks literature  \cite{gallant1990perceptron,ruder2016overview}. As noted earlier, the key  feature of a MCMC sampler  is the ability to sample a posterior probability distribution that represents the parameters of a model rather than a fixed point estimate given by optimisation methods.

\subsubsection{Likelihood}


Our Bayesian approach  for the problem requires sampling (estimating) the posterior distribution  $p(\mathbf \theta\mid\mathbf y)$ that requires the definition of both a likelihood function $p(\mathbf \theta\mid\mathbf x)$ and prior distribution $p(\theta)$. We begin  by defining the likelihood function, i.e probability of the data given the model, which is given by the product of the likelihood for every data point in the dataset of $S$ instances, as shown in Equation \ref{eqn:prod_like}

\begin{equation}
    p(\mathbf y \mid \mathbf x,\theta, \tau^2)=\prod_{t=1}^S p(\mathbf y_t\mid \mathbf x_t,\theta,\tau^2).
    \label{eqn:prod_like}
\end{equation}

We note that for our MCMC sampler, we use the log-likelihood (i.e., taking the log of the likelihood function) to eliminate numerical instabilities, which can occur since we multiple probabilities together which grows with the size of the data. It is also more convenient to maximize the log of the likelihood function since the logarithm is monotonically increasing function of its argument, i.e., maximization of the log of a function is equivalent to maximization of the function itself. In order to transform a likelihood function into a log-likelihood, we will use the log product rule as given below

\begin{equation}
    log_b(x  \times y) = log_b(x) + log_b(y).
\end{equation}

The log-likelihood simplifies the subsequent mathematical analysis and also helps avoid numerical instabilities due to the product of a large number of small probabilities. In the log-likelihood, Equation \ref{eqn:prod_like} is much simplified by computing the sum of the log probabilities as given in Equation \ref{eqn:loglhood} 

\begin{equation}
    \ln p(y \mid \mathbf x,\theta, \tau^2) =\sum_{t=1}^S\ln p(y_t\mid \mathbf x_t,\theta,\tau^2).
 \label{eqn:loglhood}
\end{equation}

In order to construct the likelihood function, we use our definition of the probability for each data point given the model as shown in Equation \ref{eqn:likey}, and the form of the Gaussian distribution as defined in Equation \ref{eqn:gauss_dist}. We use a  set of weights and biases as the model parameters $\mathbf{\theta}$ in our model $f(x, \theta)$, for $S$ training data instances and variance $\tau^2$. Our assumption of normally distributed errors leads to a likelihood given in Equation \ref{eqn:lhoodreg}

\begin{align}
    p(\mathbf y \mid \mathbf x,\theta, \tau^2) =&\notag\\ \frac{1}{(2\pi \tau ^2)^{S/2}}&\exp\left( -\frac{1}{2\tau^2}\sum_{t=1}^S \left( \mathbf y_{t}- f(\mathbf x_t, \theta)\right)^2\right).
    \label{eqn:lhoodreg}
\end{align}

\subsubsection{Prior}
\label{lin_prior}

We note that a conventional linear model transforms into a Bayesian linear model with the use of a prior distribution and a likelihood function to sample the posterior distribution via the MCMC sampler. In Section \ref{sec:basic_mcmc}, we discussed the need to define a prior distribution for our model parameters $\theta$ and $\tau^2$. In the case where the prior distribution comes from the same  family as the posterior distribution, the prior and posterior are then called conjugate distributions \cite{george1993conjugate,murphy2007conjugate}. The prior is called a \textit{conjugate prior} for the likelihood function of the Bayesian model. 

To implement conjugate priors in our linear model, we will assume a multivariate Gaussian prior for $\theta$ (Equation \ref{eqn:prior1}) and an inverse Gamma distribution (IG) for $\tau^2$ (Equation \ref{eqn:prior2}).  

\begin{equation}
 \theta \sim \mathcal{N}(0,\sigma^2)
    \label{eqn:prior1}
\end{equation}

\begin{equation}
\tau^2 \sim IG(\nu_1,\nu_2)
\label{eqn:prior2}
\end{equation}

\textcolor{black}{The noise is defined by the spread (variance) of a normal distribution. We need to define the variance that is represented by $\tau^2$ which cannot be a negative number and hence we use  IG  in Equation 18. We do not know the right value for $\tau^2$  and hence we sample this parameter in a similar fashion as $\mathbf \theta$ during the  MCMC sampling process. We note that the prior for $\tau^2$ must represent a distribution that can only sample positive real values, and we use the conjugate inverse Gamma prior with hyperparameters $\nu_1$ and $\nu_2$ representing the shape and scale parameters (see Section \ref{sec:prob_dists}).}

\textcolor{black}{We use the multivariate Gaussian distribution to represent the prior for parameters $\mathbf \theta$  such as weights and bias of the linear model, which features negative and positive real numbers. Our model features more than one parameter, hence the multivariate Gaussian distribution is most appropriate for the prior. In this example, we adopt uninformative priors with hyperparameter values of $\sigma=5$, $\nu_1=0$, and $\nu_2=0$ (Listing 6: Lines 15-17), but these values are user defined and could be refined using trial runs. These values are based on expert opinion from analysis of related trained models.}
 
First, we revisit multivariate normal distribution from Equation \ref{eqn:mvn_dist} to define the prior distribution for our linear model's parameters (weights and biases). Suppose that $ \mathbf \theta $ is our set of $M$ parameters  given by $(\mathbf \theta = \theta_1,\ldots,\theta_M)$. Since our prior is based on the normal distribution, we use mean $\mu=0$ for each parameter to ensure we sample both positive and negative real numbers. Therefore,  the mean  $ \mathbf \mu$ is a vector of zeros  and  we get the prior using 

\begin{align}
    f(\mathbf \theta) =&\notag\\ \frac{1}{\sqrt{(2\pi)^M|\mathbf\Sigma|}}
    &\exp\left(-\frac 1 2 ({\mathbf \theta} )^\mathrm{T}{\mathbf\Sigma}^{-1}({ \mathbf \theta} )\right).
 \label{eqn:fnprior}
\end{align}

The covariance matrix $\mathbf \Sigma$ is a diagonal matrix with all values equal to $\sigma^2$ (scalar). Note that ${\mathbf \Sigma}^{-1}$  becomes $I/\sigma^2$ where $I$ is an identity matrix (diagonal elements which are all ones). Hence, we take the numerator  from  Equation \ref{eqn:fnprior}, i,e.

\begin{equation}
    ({\mathbf \theta} )^\mathrm{T}{\mathbf\Sigma}^{-1} ({\mathbf \theta} )
\end{equation}

becomes

\begin{equation}
    \frac{({\mathbf \theta} )^\mathrm{T} {\mathbf I} ({\mathbf \theta} )}{\sigma^2}.
\end{equation}

We note that multiplying identity matrix with any other matrix is the matrix itself, hence finally we get $ {\mathbf \theta}^2$ in numerator. We can now move to the inverse-Gamma distribution used to define the prior for  our model's variance ($\tau^2$) and sampled (just as $\mathbf \theta$) and given by 

\begin{equation}
    f(\tau^2)=\frac{\nu_1^{\nu_2}}{\Gamma(\nu_1)}\left(\frac{1}{\tau^2}\right)^{\nu_1+1}\exp\left(\frac{-\nu_2}{\tau^2}\right).
    \label{priortau}
\end{equation}

We note that $\nu_1^{\nu_2}/\Gamma(\nu_1)$ is a constant which can be dropped considering proportionality.  We take into account the product of all our sampled parameters ($\theta$ and $\tau^2$) to define the combined prior, as given by Equation \ref{eqn:prior} 

\begin{align}
p(\mathbf{\theta}) \propto& \frac{1}{(2\pi\sigma^2)^{M/2}}\times\notag\\
&\exp\Bigg\{-\frac{1}{2\sigma^2}\bigg( \sum_{i=1}^M \mathbf \theta^2 \bigg) \Bigg\}
\times\notag\\
&\tau^{-2(1+\nu_1)}\exp\left(\frac{-\nu_2}{\tau^2}\right).
\label{eqn:prior} 
\end{align}

\subsubsection{Python Implementation}

\textcolor{black}{The  Python code presented in Listing \ref{lst:linear-reg-model}\footnote{\href{https://github.com/sydney-machine-learning/Bayesianneuralnetworks-MCMC-tutorial/blob/main/03-Linear-Model.ipynb}{https://github.com/sydney-machine-learning/Bayesianneuralnetworks-MCMC-tutorial/blob/main/03-Linear-Model.ipynb}} begins the implementation of a Bayesian linear model using MCMC sampling. First, we define our simple linear model as given in Equation \ref{eqn:linear_model} using class LinearModel (Line 1) of  Listing \ref{lst:linear-reg-model} and define functions to evaluate the proposal (line 13), get the prediction (Line 25) and encode the parameters proposed from the MCMC sampler class into the linear model (Line 30).}

\begin{figure*}[htbp!]
\begin{lstlisting}[language=Python, label={lst:linear-reg-model}, caption=Python implementation of a simple linear regression model]
class LinearModel:
    '''
    Simple linear model with a single output (y) given the covariates x_1...x_M of the form:
    y = w_1 * x_1 + ... + w_M * x_M + b
    where M = number of features, w are the weights, and b is the bias.
    '''
    # Initialise values of model parameters
    def __init__(self):
        self.w = None
        self.b = None 

    # Function to take in data and parameter sample and return the prediction
    def evaluate_proposal(self, data, theta):
        '''
        Encode the proposed parameters and then use the model to predict
        Input:
            data: (N x M) array of data
            theta: (M + 1) vector of parameters. The last element of theta consitutes the bias term (giving M + 1 elements)
        '''
        self.encode(theta)  # method to encode w and b
        prediction = self.predict(data) # predict and return
        return prediction

    # Linear model prediction
    def predict(self, x_in):
        y_out = x_in.dot(self.w) + self.b 
        return y_out
    
    # Helper function to split the parameter vector into w and band store in the model
    def encode(self, theta):
        self.w =  theta[0:-1]
        self.b = theta[-1]
\end{lstlisting}
\end{figure*}

Now that we have a class for our linear model, we can define the functions that will allow us to carry out MCMC sampling for the model parameters. We  define our log-likelihood function  using Equation \ref{eqn:lhoodreg}, which becomes

\begin{align}
    logp(\mathbf y \mid \mathbf x,\mathbf \theta, \tau^2) =&\notag\\ - log((2\pi \tau ^2)^{S/2})& -\frac{1}{2\tau^2}\sum_{t=1}^S \left( \mathbf y_{t}- f( \mathbf x_t, \mathbf \theta)\right)^2.
    \label{eqn:log_lhoodreg}
\end{align}

\textcolor{black}{Furthermore, we define our log-prior using Equation \ref{eqn:prior} which becomes}

\begin{align}
    \log{p(\mathbf{\mathbf \theta})} \propto& -\frac{M}{2}\log{2\pi\sigma^2}\notag\\ 
    &-\frac{1}{2\sigma^2}\bigg( \sum_{i=1}^M \mathbf \theta^2 \bigg)
    \notag\\ &-(1+\nu_1)\log{\tau^{2}}-\frac{\nu_2}{\tau^2}.
\label{eqn:log_prior}
\end{align}



\textcolor{black}{We present Python implementation of the log-likelihood (Lines 2 - 18) and the prior (Lines 21 - 37) in Listing \ref{lst:gaussian-likelihood-prior}.}

\begin{figure*}[htbp!]
\begin{lstlisting}[language=Python, label={lst:gaussian-likelihood-prior}, caption=Python implementation of likelihood and prior functions for linear regression model to be incorporated into the MCMC sampling class]
# Define the log-likelihood function
def likelihood_function(self, theta, tausq):
    '''
    Calculate the likelihood of the data given the parameters
    Input:
        theta: (M + 1) vector of parameters. The last element of theta consitutes the bias term (giving M + 1 elements)
        tausq: variance of the error term
    Output:
        log_likelihood: log likelihood of the data given the parameters
        model_prediction: prediction of the model given the parameters
        accuracy: accuracy (RMSE) of the model given the parameters
    '''
    # first make a prediction with parameters theta
    model_prediction = self.model.evaluate_proposal(self.x_data, theta) 
    accuracy = self.rmse(model_prediction, self.y_data) #RMSE error metric 
    # now calculate the log likelihood
    log_likelihood = np.sum(-0.5 * np.log(2 * np.pi * tausq) - 0.5 * np.square(self.y_data - model_prediction) / tausq)
    return [log_likelihood, model_prediction, accuracy] 

# Define the prior
def prior(self, sigma_squared, nu_1, nu_2, theta, tausq): 
    '''
    Calculate the prior of the parameters
    Input:
        sigma_squared: variance of normal prior for theta
        nu_1: parameter nu_1 of the inverse gamma prior for tau^2
        nu_2: parameter nu_2 of the inverse gamma prior for tau^2
        theta: (M + 1) vector of parameters. The last element of theta consitutes the bias term (giving M + 1 elements)
        tausq: variance of the error term
    Output:
        log_prior: log prior
    '''
    n_params = self.theta_size # number of parameters in model
    part1 = -1 * (n_params / 2) * np.log(sigma_squared)
    part2 = 1 / (2 * sigma_squared) * (sum(np.square(theta)))
    log_prior = part1 - part2 - (1 + nu_1) * np.log(tausq) - (nu_2 / tausq)
    return log_prior
\end{lstlisting}
\end{figure*}

\textcolor{black}{Before running the MCMC sampler, we need to set up the sampler hyperparameters, such as the maximum sampling time and burn-in period (Listing }\ref{lst:linear-sampling}). We also need to assign hyperparameters that define the priors such as Gaussian prior variance ($\sigma^2$) and the IG prior parameters, $\nu_1$ and $\nu_2$ (Listing \ref{lst:linear-mcmc-class}: Lines 15, 16 and 17). First, we need to generate an initial sample for our parameters and initialise arrays to capture the samples that form the posterior distribution, the accuracy, and the model predictions as shown in code Listing \ref{lst:linear-mcmc-sampler} (Lines 5-23). Then we proceed with sampling as per the MCMC sampling algorithm detailed in Algorithm \ref{alg:simple_mcmc} and code Listing \ref{lst:linear-mcmc-sampler}. This algorithm uses a Gaussian random walk for the parameter proposals ($\theta_p$ and $\tau^2_p$), perturbing the previous proposed value with Gaussian noise as shown in Equations \ref{eqn:theta} and \ref{eqn:tau}, respectively.

\begin{equation}
\mathbf \theta_p \sim \mathbf \theta_{p-1} + \mathcal{N}(0,  \Delta_{\theta})
\label{eqn:theta}
\end{equation}

\begin{equation}
 \eta_p \sim \eta_{p-1} + \mathcal{N}(0, \Delta_{\eta})
 \label{eqn:tau}
\end{equation}

We implement the MCMC sampler with a Gaussian random-walk proposal for $\eta_p = \log \tau_p^2$,  where we use $\eta$ to represent $\tau^2$ in log-space (Listing 5: Line 29). The step sizes for the proposals are determined by the hyperparameters $\Delta_{\theta}$ and $\Delta_{\eta}$ which define the variance for the proposal of $\theta_p$ and $\eta_p$ respectively. Once we sample $\eta$, we take the exponential to convert it back to the original form (see Line 30) and obtain $\tau^2$.

After getting the proposal for the parameters i.e. $\mathbf \theta$ and $\tau^2$, we  call the log-likelihood and prior functions to obtain their respective values, as shown in Lines 32 - 35 of Listing \ref{lst:linear-mcmc-sampler}. 
Note that the log-likelihood is used and hence the ratio of previous and current likelihood will need to consider log laws (rules), i.e., we note the log product rule in Equation \ref{eqn:prod} and the quotient rule in Equation \ref{eqn:quo}. \textcolor{black}{We use these rules in Lines 37 and 38 of Listing \ref{lst:linear-mcmc-sampler}. Based on  Equation \ref{eqn:cond} and taking the quotient rule into account since we are in the log space, we  then accept/reject the proposed value according to the Metropolis-Hastings acceptance ratio (Line 42)  as shown in  Lines 41 - 54.}
 
\begin{equation}
log_b(x  \times y) = log_b(x) + log_b(y)
 \label{eqn:prod}
\end{equation}

\begin{equation}
log_b(x / y) = log_b(x) - log_b(y)
\label{eqn:quo}
\end{equation}
\begin{figure*}[htbp!]
\begin{lstlisting}[language=Python, label={lst:linear-mcmc-sampler}, caption=Python implementation of an MCMC sampler for the linear model]
# MCMC sampler
def sampler(self):
    # Run the sampler for a defined linear model
    # Define empty arrays to store the sampled posterior values
    pos_theta = np.ones((self.n_samples, self.theta_size)) 
    # posterior defining the variance of the noise in predictions
    pos_tau = np.ones((self.n_samples, 1))
    # record output f(x) over all samples
    pred_y = np.ones((self.n_samples, self.x_data.shape[0]))
    # record the RMSE of each sample
    rmse_data = np.zeros(self.n_samples)

    ## Initialisation:initialise theta - the model parameters
    theta = np.random.randn(self.theta_size)
    # make initial prediction
    pred_y[0,] = self.model.evaluate_proposal(self.x_data, theta)
    # initialise eta - we sample eta as a gaussian random walk in the log space of tau^2
    eta = np.log(np.var(pred_y[0,] - self.y_data))
    tausq_proposal = np.exp(eta)
    # calculate the prior
    prior_val = self.prior(self.sigma_squared, self.nu_1, self.nu_2, theta, tausq_proposal)
    # calculate the likelihood considering observations
    [likelihood, pred_y[0,], _, rmse_data[0]] = self.likelihood_function(theta, tausq_proposal)

    ## Run the MCMC sample for n_samples
    for ii in np.arange(1,self.n_samples):
        # Sample new values for theta and tau using a Gaussian random walk
        theta_proposal = theta + np.random.normal(0, self.step_theta, self.theta_size)
        eta_proposal = eta + np.random.normal(0, self.step_eta, 1) # sample tau^2 in log space
        tausq_proposal = np.exp(eta_proposal)
        # calculate the prior
        prior_proposal = self.prior(
        self.sigma_squared, self.nu_1, self.nu_2, theta_proposal, tausq_proposal)
        # calculate the log-likelihood considering observations
        [likelihood_proposal, pred_y[ii,], _, rmse_data[ii]] = self.likelihood_function(theta_proposal,tausq_proposal)
        # Noting that likelihood_function and prior_val return log likelihoods, we can calculate the acceptance probability
        diff_likelihood = likelihood_proposal - likelihood
        diff_priorlikelihood = prior_proposal - prior_val
        mh_prob = min(1, np.exp(diff_likelihood + diff_priorlikelihood))
        # sample to accept or reject the proposal according to the acceptance probability
        u = np.random.uniform(0, 1)
        if u < mh_prob:
            # accept and update the values
            likelihood = likelihood_proposal
            prior_val = prior_proposal
            theta = theta_proposal
            eta = eta_proposal
            # store to make up the posterior
            pos_theta[ii,] = theta_proposal
            pos_tau[ii,] = tausq_proposal
        else:
            # reject move and store the old values
            pos_theta[ii,] = pos_theta[ii-1,]
            pos_tau[ii,] = pos_tau[ii-1,]
    # store the posterior (samples after burn in) in a pandas dataframe and return
    self.pos_theta = pos_theta[self.n_burnin:, ]
    self.pos_tau = pos_tau[self.n_burnin:, ] 
    self.rmse_data = rmse_data[self.n_burnin:]
    # split theta into w and b
    results_dict = {'w{}'.format(_): self.pos_theta[:, _].squeeze() for _ in range(self.theta_size-1)}
    results_dict['b'] = self.pos_theta[:, -1].squeeze()
    results_dict['tau'] = self.pos_tau.squeeze()
    results_dict['rmse'] = self.rmse_data.squeeze()
    results_df = pd.DataFrame.from_dict(results_dict)
    return results_df
\end{lstlisting}
\end{figure*}

Now that we have the sampler code (Listing \ref{lst:linear-mcmc-sampler}), we can create an MCMC class that brings together the model, data, hyperparameters and sampling algorithm as shown in Listing \ref{lst:linear-mcmc-class}.

\begin{figure*}[htbp!]
\begin{lstlisting}[language=Python, label={lst:linear-mcmc-class}, caption=Python implementation of an MCMC class for Bayesian linear model.]
class MCMC:
    def __init__(self, n_samples, n_burnin, x_data, y_data):
        self.n_samples = n_samples # number of MCMC samples
        self.n_burnin = n_burnin # number of burn-in samples
        self.x_data = x_data # (N x M)
        self.y_data = y_data # (N x 1)
        self.theta_size = x_data.shape[1] + 1 # weights for each feature and a bias term (M+1)

        # MCMC sampler hyperparameters - defines the variance term in our Gaussian random walk
        self.step_theta = 0.02;  
        self.step_eta = 0.01; # note eta is used as tau in the sampler to consider log scale.  
        
        # model hyperparameters
        # considered by looking at distribution of  similar trained  models - i.e distribution of weights and bias
        self.sigma_squared = 5
        self.nu_1 = 0
        self.nu_2 = 0

        # initisalise the linear model class
        self.model = LinearModel()

        # store output
        self.pos_theta = None
        self.pos_tau = None
        self.rmse_data = None

        # functions defined above - this is poor practice, but done for readability 
        # and clarity
        self.likelihood_function = MethodType(likelihood_function, self)
        self.prior_val = MethodType(prior_val, self)
        self.sampler = MethodType(sampler, self)

    def model_draws(self, num_samples = 10):
        '''
        Simulate new model predictions (mu) under the assumption that our posteriors are 
        Gaussian.
        '''
        # num_samples x num_data_points
        pred_y = np.zeros((num_samples,self.x_data.shape[0]))
        sim_y = np.zeros((num_samples,self.x_data.shape[0]))

        for ii in range(num_samples):
            theta_drawn = np.random.normal(self.pos_theta.mean(axis=0), self.pos_theta.std(axis=0), self.theta_size)
            tausq_drawn = np.random.normal(self.pos_tau.mean(), self.pos_tau.std())

            [_, pred_y[ii,:], sim_y[ii,:],_] = self.likelihood_function(
                theta_drawn, tausq_drawn
            )
        return pred_y, sim_y
\end{lstlisting}
\end{figure*}

\begin{figure*}[htbp!]
\begin{lstlisting}[language=Python, label={lst:linear-sampling}, caption=Code to call the MCMC sampling class and fit a model to some toy data]
## MCMC Settings and Setup
n_samples = 20000 # number of samples to draw from the posterior
burn_in = int(n_samples* 0.25) # number of samples to discard before recording draws from the posterior

# Generate toy data
n_data = 100
n_features = 1
x_data = np.repeat(np.expand_dims(np.linspace(0, 1, n_data),axis=-1),n_features,axis=1)
y_data = 3 * x_data[:,0] + 4 + np.random.randn(n_data) * 0.5

# Initialise the MCMC class
mcmc = MCMC(n_samples, burn_in, x_data, y_data)
# Run the sampler
results = mcmc.sampler()
# Draw sample models from the posterior
pred_y, _ = mcmc.model_draws(num_samples=100)
\end{lstlisting}
\end{figure*}

 We can then run the MCMC sampler function as shown in Listing \ref{lst:linear-sampling} Line 12. After the code runs, we get the results (Line 14) and can generate the predictions from model posterior draws (Line 16) of the trained Bayesian linear model.
 
\subsection{Bayesian neural networks via MCMC}
\subsubsection{Neural networks}

We utilise a simple neural network, also known as a multilayer perceptron to demonstrate the process of training a Bayesian neural network via the MCMC sampler. A neural network model $f(x)$ is made up of a series of lower level computations, which can be used to transform inputs to their corresponding outputs $\{\bf \bar{x}_t,\bf{y}_t\}$. Neural networks feature layers of neurons whose value is determined based on a linear combination of inputs from the previous layer, with an activation function used to introduce nonlinearity. 

\begin{figure}[!htb]
    \centering
    \begin{tikzpicture}[
        node distance=3cm,
        neuron/.style={circle, draw, minimum size=0.7cm},
        weight/.style={minimum size=0.6cm, align=center, font=\footnotesize},
        layer/.style={align=center, font=\footnotesize}
    ]
    
    \foreach \i in {1,...,4} {
        \node[neuron] (I-\i) at (0,-\i) {};
    }
    \node[layer, above of=I-1, node distance=1cm] {Inputs\\$(\bar{\mathbf{x}_t})$};
    
    \foreach \i in {1,...,5} {
        \node[neuron] (H-\i) at (3,-\i+0.5) {};
    }
    \node[layer, above of=H-1, node distance=1cm] {Hidden\\neurons};
    
    \node[neuron, right of=H-3, node distance=3cm] (O) {};
    \node[layer, above of=O, node distance=1cm] {Output\\$(\mathbf{y}_t)$};
    
    \node[weight] (WI) at (1.5,0) {Weights\\ $\left(\tilde{\mathbf{w}},\delta_h\right)$};
    \node[weight] (WH) at (4.5,-0.6) {Weights\\ $\left(\tilde{\mathbf{v}},\delta_o\right)$};
    
    \foreach \i in {1,...,4} {
        \foreach \j in {1,...,5} {
            \draw (I-\i) -- (H-\j);
        }
    }
    \foreach \i in {1,...,5} {
        \draw (H-\i) -- (O);
    }
    
    \end{tikzpicture}
    \caption{Simple neural network with a single hidden layer. The information is passed and processed from the input to hidden and then finally to the output layer. }
    \label{neuralnetworkpic}
\end{figure}
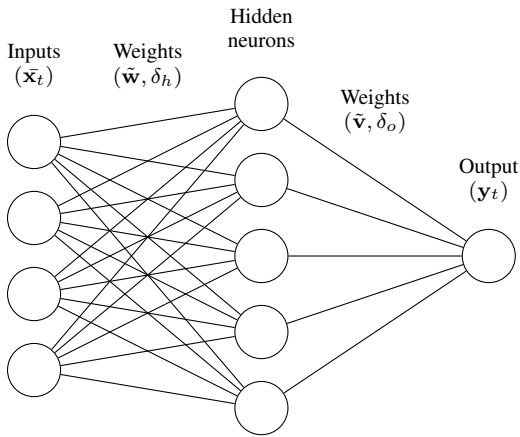

We consider a simple neural network with one hidden layer with four input neurons, five hidden neurons and one output neuron, as shown in Figure \ref{neuralnetworkpic}. As an example, we can calculate the output value of the $j$th neuron in the first hidden layer of a network ($h,j$) using a weighted combination of the $m$ inputs ($\bar{x}_{t}$), as shown in Equation \ref{eqn:nn}.

\begin{equation}
g\bigg(\delta_{h,j} + \sum_{i=1}^{m} w_{i,j} 
    \bar{x}_{t,i} \bigg)
\label{eqn:nn}
\end{equation}

where, the bias ($\delta_{h,j}$) and weights ($w_i$ for each of the $m$ inputs) are parameters to be sampled (trained or estimated), and $g(.)$ is the activation function that is used to perform a nonlinear transformation. In our case, the function $g(.)$ is the \textit{sigmoid activation function}, used for the hidden and output layers  as shown in Figure \ref{neuralnetworkpic}.

We train the model to approximate the function $f$ such that $f(\mathbf x)= \bar{y}$ for all input-output pairs of instances from the training dataset. We extend our previous calculation, to a single neuron in the hidden layer to calculate the output $f(\bf \bar{x}_t)$ as shown in Equation \ref{eqn:fx}.

\begin{equation}
    f(\mathbf x)  =  g \bigg(  \delta_o + 
    \sum_{j=1}^{H} v_{h} \times g \bigg(  \delta_h + \sum_{i=1}^{m} w_{i,j} 
    \mathbf x_{i} \bigg)\bigg)
 \label{eqn:fx}
\end{equation}

where $H$ is the number of neurons in the hidden layer, $\delta_o$ is the bias for the output, and $v_h$ are the weights from the hidden layer to the output neuron. The complete set of parameters for the neural network model (Figure \ref{neuralnetworkpic}) is made up of $\theta=(\mathbf {\tilde{w}},\mathbf {\tilde {v}},{\mathbf \delta})$, where ${\mathbf \delta}=({\mathbf \delta_o},{\mathbf \delta_h})$. $\mathbf {\tilde{w}}$ are the weights transforming the input to hidden layer.  $\mathbf {\tilde {v}}$ are the weights transforming the hidden to output layer. ${\mathbf \delta_h}$ \textcolor{black}{is the bias for  the hidden layer, and  ${\mathbf \delta_o}$ is the bias for  the output layer. }

\subsubsection{Bayesian neural networks}

\begin{figure}[htbp!]
   \centering 
\includegraphics[width=8cm]{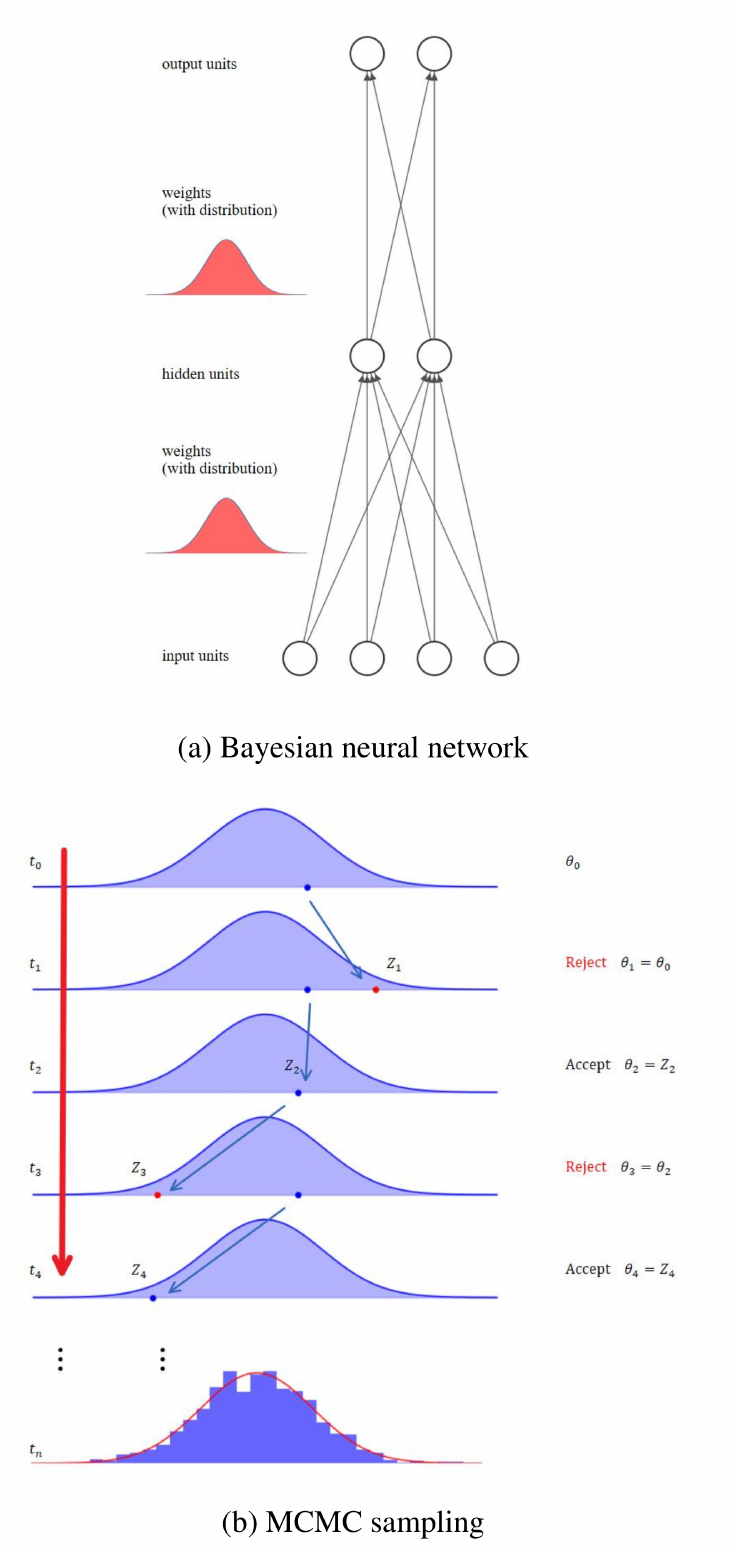}
  \caption{Bayesian neural network and MCMC sampling adapted from \cite{chandra_bayesian_2021}.}
   \label{fig:neuralnetworkpic}
 \end{figure}

A Bayesian neural network is a probabilistic implementation of a standard neural network with the key difference being that the weights and biases are represented via the posterior probability distributions rather than single point values as shown in Figure \ref{fig:neuralnetworkpic}. Similar to canonical neural networks \cite{hornik1989multilayer}, Bayesian neural networks also have universal continuous function approximation capabilities. However, the posterior distribution of the network parameters allows uncertainty quantification on the predictions.

The task for MCMC sampling is to estimate (sample) the posterior distributions representing the weights and biases of the neural network that best fit the data. Perhaps, it can be argued that the method should be called an estimator, but we will stick to the sampler as given in the literature. As in the previous examples, we begin inference with prior distributions over the weights and biases of the network and use a sampling scheme to find the posterior distributions given training data. Since non-linear activation functions exist in the network, the conjugacy of prior and posterior is lost. Therefore, we must employ an MCMC sampling scheme and make assumptions about the distribution of errors.

We specify the model similar to the Bayesian linear regression, assuming a Gaussian error as given in Equation \ref{eqn:error}.

\begin{equation}
    y=f(\mathbf x,\theta)+e \qquad e\sim \mathcal{N}(0,\tau^2)
 \label{eqn:error}
\end{equation}

This leads to the same likelihood function as presented in logarithmic form in Equation \ref{eqn:log_lhoodreg}. As in Section \ref{lin_prior}, we adopt Gaussian priors for all parameters of the model ($\theta$), with zero mean and a user-defined variance ($\sigma^2$), and an IG distribution for the variance of the error model ($\tau^2$), with parameters $\nu_1$ and $\nu_2$. The \textit{likelihood function} and \textit{prior} function, therefore remain unchanged from their definition in Listing \ref{lst:gaussian-likelihood-prior}.

\subsubsection{Multinomial likelihood for classification problems}

\textcolor{black}{We note that neural network models are also prominent for classification problems apart from regression and prediction problems. Bayesian neural networks via the MCMC sampler require an appropriate likelihood function that is more suitable for discrete data to capture classification problems. Hence, we use the multinomial likelihood, which is applicable to both binary and multi-class classification problems. Hence,  we define  the multinomial log-likelihood function for the classification problems using}

\begin{eqnarray}
\log\left(p({\bf y }|\mathbf{\theta})\right)&=&\sum_{t\in N}\sum_{k=1}^K z_{t,k}\log{\pi_k}
\label{multinomial}
\end{eqnarray}
\noindent \textcolor{black}{for classes $ k = 1,\ldots,K$, where $\pi_k$ is the output  of the neural network  model after applying the transfer function, and $N$ is the number of instances in the training data.  In this case, we utilize the \textit{softmax function} \cite{bishop1995neural} as the transfer function:}
\begin{equation}
\pi_k = \frac{\exp(f(x_p))}{\sum_{k=1}^K \exp(f(x_k))}
\end{equation}
\noindent for $k = 1,\ldots,K$. $z_{t,k}$ is an indicator variable for the given instance of data $t$. We define class $k$  in the data by  
\begin{equation}
z_{t,k} =
\begin{cases}
    1,& \text{if } y_t = k\\
    0,              & \text{otherwise.}
\end{cases}
\end{equation}

\textcolor{black}{We note that we do not use the noise parameter (i.e., $\tau^2$) as in the case of the inverse gamma distribution for the Gaussian likelihood earlier, hence we do not need to have a prior distribution for this case. We will use a Gaussian prior for weights and biases of the neural network model. Therefore, in the case of classification, our prior distribution from Equation \ref{eqn:log_prior} simplifies to}

\begin{align}
    p(\mathbf{\theta}) \propto& \frac{1}{(2\pi\sigma^2)^{M/2}}\times\notag\\ 
    &\exp\Bigg\{-\frac{1}{2\sigma^2}\bigg( \sum_{i=1}^M \mathbf \theta^2 \bigg) \Bigg\}.
\label{eqn:priorclass}
\end{align}

\textcolor{black}{and finally log-prior becomes }

\begin{align}
    \log{p(\mathbf{\theta})} \propto& -\frac{M}{2}\log{2\pi\sigma^2}\times\notag\\ 
    &-\frac{1}{2\sigma^2}\bigg( \sum_{i=1}^M \mathbf \theta^2 \bigg) 
\label{eqn:log_priorclass}
\end{align}

\subsubsection{Training neural networks via backpropagation}

We note that typically random-walk proposal distributions are used for small scale-models such as linear models; however, neural network models features a large number of parameters. The choice of a proposal distribution is essential for models with large number of parameters. We need to incorporate gradients into our proposal distribution for better sampling,  and we will begin by examining how gradients are incorporated in conventional neural networks and deep learning models.

Gradient-based methods have been widely used in machine  learning  and served as the backbone in the backpropagation algorithm  \cite{goodfellow2016deep}. A prominent implementation is \textit{stochastic gradient descent} (SGD) which involves stepping through the parameter space iteratively in a stochastic manner using gradients, to optimize a differentiable objective function. The method has been prominently featured in the backpropagation algorithm for training of various neural networks architectures including deep learning models \cite{rumelhart1986learning,wythoff1993backpropagation,rumelhart1995backpropagation}. Backpropagation involves a forward pass which propagates information forward to get prediction (decision) at the output layer and a backward pass to compute the local gradients for each of the parameters. These gradients are then used to inform the update of the model parameters in an interative process, where  the weights and biases are updated at each step.  The training of neural networks is also considered as solving a non-convex optimization problem $\argmin{L(\theta)}$; where, $\theta \in R^n$ is the set of parameters and $L$ is the loss function. We give the parameter (weight) update  for an iteration (epoch)  of SGD in Equation \ref{eqn:bp}

\begin{equation}
\theta_k = \theta_{k-1} - a_{k-1} \nabla L(\theta_{k-1})
\label{eqn:bp}
\end{equation}
where, $\theta_{k}$ denotes the $k^{th}$ iteration, $a_k$ is the learning rate, and $ \nabla L(\theta_k) $ denotes the  gradient.


We note that the \textit{learning rate} is user defined hyperparameter which depends on the problem and data at hand. It is typically determined through tuning using cross-validation or trial and error. A number of extensions of the backgropagation algorithm employing SGD have been proposed to address limitations. These include use of \textit{weight decay} regularization during training to improve generalisation ability \cite{krogh1991simple}, a momentum mechanism for faster  training \cite{sutskever2013importance,moreira1995neural},  adaptive learning rate \cite{moreira1995neural}, and second-order gradient methods \cite{battiti1992first} which, although efficient, have problems in scaling up computationally with larger models. In the last decade, with the deep learning revolution, further attempts have been made to ensure that enhanced backpropagation algorithms, not only improve training accuracy, but can also scale well computationally for large deep learning models. These led to the development of methods such as the adaptive gradient algorithm (AdaGrad) \cite{duchi2011adaptive}, Ada-Delta \cite{zeiler2012adadelta}, and Adam (adaptive moment estimation) \cite{kingma_2015}. These advanced algorithms  are generally based on the idea of adapting the learning rate automatically during the training, taking into account the recent history of the optimisation process. 


\paragraph{Langevin proposal distribution}

We mentioned earlier that random-walk proposal distributions are suited for small scale models and better proposal distributions would be required for neural network models. Although simple neural networks have much lower number of parameters, when compared to deep learning models, training simple neural networks with MCMC sampling is a challenge with random-walk proposal distribution. We need to utilise the properties of backpropagation algorithm and the mechanism of weight update using gradients. 
Hence,  we utilise \textit{stochastic gradient Langevin dynamics} \cite{welling2011bayesian} for the proposal distribution which features the addition of noise to the stochastic gradients. The method has shown to be effective for linear models \cite{welling2011bayesian} which motivated its use in Bayesian neural networks. In the literature, Langevin MCMC has been very promising for simple and deep neural networks \cite{Chandra2019NC,chandra2021bayesian,chandra2021revisiting}. Hence, we draw the proposed values for the parameters ($\mathbf\theta^p$)  according to a one-step (epoch) gradient as shown in Equation \ref{eqn:update}.

 \begin{equation}
\mathbf\theta^p \sim \mathcal{N}(\bar{\mathbf\theta}^{[s]},\Sigma_{\theta})
\label{eqn:update} 
 \end{equation}

A Gaussian distribution with a standard deviation of $\Sigma_{\theta}$, and mean ($\bar{\mathbf\theta}^{[s]}$) calculated using a gradient based update (Equation  \ref{gradient} of the parameter values from the previous step (${\mathbf\theta}^{[s]}$).

 \begin{equation}
    \bar{\mathbf\theta}^{[s]}=\mathbf\theta^{[s]} +r\times\nabla  E( \mathbf{\theta^{[s]}})
 \label{gradient} 
 \end{equation}
 
with learning rate $r$ and gradient update ($\nabla E(\mathbf\theta^{[s]})$) according to the model residuals.
 \begin{equation}
    E(\mathbf\theta^{[s]}) = \sum_{t\in{\mathcal{T}}}(y_t- F({\bf x}_i, \mathbf{\theta^{[s]}}))^2
  \label{eqn:resi} 
 \end{equation}
 
 \begin{equation}
    \nabla E(\mathbf\theta^{[s]})= 
    \left(\frac{\partial{E}}{\partial{\theta_1}},\ldots, 
    \frac{\partial{E}}{\partial{\theta_{L}}}\right).\nonumber
 \end{equation}

Hence, the Langevin proposal distribution (also known as Langevin-gradient) consists of 2 parts: 
\begin{enumerate}
    \item  Gradient descent-based weight update  
    \item Addition of Gaussian noise from $\mathcal{N}(0,\Sigma_{\theta})$ 
\end{enumerate}

We need to ensure that the detailed balance is maintained since the Langevin-gradient proposals are not symmetric. We note that MCMC implementations with relaxed detailed balance conditions for some applications also exist \cite{suwa2010markov}.  Therefore, we use a combined update in the Metropolis-Hastings step, which accepts the proposal  $\mathbf\theta^p$ for a position $s$ with the probability $\alpha$, as shown in Equation \ref{met_rat}

\begin{equation}
    \alpha=\min\left\{1,\frac{p(\mathbf\theta^p|\mathbf {y})q(\mathbf\theta^{[s]}|\mathbf\theta^p)}{p(\mathbf\theta^{[s]}|\mathbf {y})q(\mathbf\theta^p|\mathbf\theta^{[s]})}\right\}
    \label{met_rat} 
\end{equation}

where $p(\mathbf\theta^p|\mathbf {y})$ and $p(\mathbf\theta^{[s]}|\mathbf {y})$  can be computed using the likelihood and prior (Equations \eqref{eqn:lhoodreg} and \eqref{eqn:prior}). We give the ratio of the proposed and the current $q(\mathbf\theta^p|\mathbf\theta^{[s]})$ in Equation \ref{eqn:lg} 

\begin{equation}
q(\mathbf\theta^{[s]}|\mathbf\theta^{p}) \sim N(\bar{\mathbf\theta}^{[s]},\Sigma_{\mathbf\theta})
\label{eqn:lg}
\end{equation}

which is based on a one-step (epoch) gradient  $ \nabla E_{\bf y}[\mathbf\theta^{[s]}$ and learning rate $r$, as given in Equation \ref{eqn:gradient}

\begin{equation}
\bar{\mathbf\theta}^{[s]}=\mathbf\theta^{[s]} +r\times\nabla E_{\bf y}[\mathbf\theta^{[s]} ].
\label{eqn:gradient}
\end{equation}

Thus, this ensures that the detailed balance condition holds and the sequence  ${\mathbf\theta}^{[s]}$ converges to draws from the posterior $p(\mathbf\theta|\bf y)$.  Since our implementation is in the log-scale,   we give the log-posterior   in Equation  \ref{eqn:logpos}

\begin{align}
 \log\left(p(\mathbf{\theta}|{\bf y})\right)=&\notag\\\log\left(p(\mathbf{\theta})\right)&+\log\left(p({\bf y}|\mathbf{\theta})\right) + \log( q(\mathbf{\theta}|\mathbf{\theta^*})) 
 \label{eqn:logpos}
\end{align}


Algorithm \ref{alg:lvbnn} gives a full description of the Langevin MCMC sampling scheme with user-defined parameters that include the maximum number of samples ($S_{max}$), rate of Langevin-gradient proposals ($L_{prob}$), and learning rate  $r$ used for the Langevin-gradient proposals. We note that in a standard Langevin MCMC approach,  $L_{prob}=1$ and Gaussian noise is already part of Langevin-gradient distribution. However, in our implementation, we use a combination of random-walk proposal distribution with Langevin-gradients, as this is computationally more efficient than Langevin-gradients alone, which consumes extra computational time when computing gradients, especially in larger models. 

We begin by drawing initial values for the $\theta$ from the prior distribution given in Equation \eqref{eqn:prior} (Stage 1.1). We draw a new proposal  for $\theta^p$ (which incorporates the model weights and biases and $\tau^2$ from either a Langevin-gradient or random-walk proposal distribution (Stage 1.2). We evaluate the proposal  using the Bayesian neural network (BNN) model with the log-likelihood function in Equation \ref{eqn:lhoodreg} (Stage 1.4) and the prior given in Equation \eqref{eqn:prior} (Stage 1.3).  We can then check if the proposal should be accepted using Metropolis-Hastings condition (Stage 1.5 and 1.6). If accepted, the proposal becomes part of the chain, else we retain previous the last accepted state of the chain. We repeat the procedure until the maximum samples are reached ($S_{max}$). Finally, we execute post-sampling stage, where we obtain the posterior distribution by concatenating the history of the samples in the chain.

\begin{algorithm} 
\KwData{Dataset}
\KwResult{Posterior distribution of  model parameters (weights and biases)} 

- Stage 1.0: Metropolis Transition\\
- 1.1 Draw initial values $\theta_{0}$ from the prior \\
\For{each  $s$ until $S_{max}$}{
    
    1.2 Draw $\kappa$ from a Uniform-distribution [0,1] \\
    \If{$\kappa  \leq L_{prob}$}{ 
        Use Langevin-gradient proposal distribution: $\mathbf\theta^p \sim \mathcal{N}(\bar{\mathbf\theta}^{[s]},\Sigma_{\theta})$\\} 
    \Else{   
        Use random-walk proposal distribution: $\mathbf\theta^p \sim \mathcal{N}(\mathbf\theta^{[s]},\Sigma_{\theta})$\\}
    1.3 Evaluate prior given in Equation \ref{eqn:prior} \\
    1.4 Evaluate log-likelihood given in Equation \ref{eqn:lhoodreg}  \\
1.5 Compute the posterior probability  for Metropolis-Hastings condition -  Equation \ref{eqn:logpos}\\

    1.6 Draw $u$ from a Uniform-distribution [0,1]  \\
    \If{$\log(u) \leq log(p({\bf y }|\mathbf{\theta}))$}{ %
        Accept  replica state: $\theta^{[s+1]} \leftarrow \theta^p$
            }
    \Else{
        Reject and retain previous state: $\theta^{[s+1]} \leftarrow \theta^{[s]}$
  }
    }

\caption{Bayesian neural network  via Langevin MCMC sampling.}
\label{alg:lvbnn}
\end{algorithm}


\begin{figure*}[htbp]
\begin{lstlisting}[language=Python, label={lst:nn-model}, caption=Python implementation of the Neural Network class]
class NeuralNetwork:
    '''
    Neural Network model with a single hidden layer and a single output (y)
    '''
    def __init__(self, layer_sizes,learning_rate=0.01):
        '''  Initialize the model
        Input:
            - layer_sizes (input, hidden, output): array specifying the number of 
            nodes in each layer
            - learning_rate: learning rate for the gradient update
        '''
        # Initial values of model parameters
        self.input_num = layer_sizes[0]
        self.hidden_num = layer_sizes[1]
        self.output_num = layer_sizes[2]
        # total number of parameters from weights and biases
        self.n_params = (self.input_num * self.hidden_num) + (self.hidden_num * self.output_num) +\
        self.hidden_num + self.output_num
        # learning params
        self.lrate = learning_rate
        # Initialize network structure
        self.initialise_network()
        # functions defined above - this is poor practice, but done for readability 
        # and clarity
        self.forward_pass = MethodType(forward_pass, self)
        self.backward_pass = MethodType(backward_pass, self)
    
    def initialise_network(self):
        ''' 
        Initialize network structure - weights and biases for the hidden layer and output layer
        '''
        # hidden layer
        self.l1_weights = np.random.normal(
            loc=0, scale=1/np.sqrt(self.input_num),
            size=(self.input_num, self.hidden_num))
        self.l1_biases = np.random.normal(
            loc=0, scale=1/np.sqrt(self.hidden_num), 
            size=(self.hidden_num,))
        # placeholder for storing the hidden layer values
        self.l1_output = np.zeros((1, self.hidden_num))
        # output layer
        self.l2_weights = np.random.normal(
            loc=0, scale=1/np.sqrt(self.hidden_num), 
            size=(self.hidden_num, self.output_num))
        self.l2_biases = np.random.normal(
            loc=0, scale=1/np.sqrt(self.hidden_num), 
            size=(self.output_num,))
        # placeholder for storing the model outputs
        self.l2_output = np.zeros((1, self.output_num))

    def evaluate_proposal(self, x_data, theta):
        '''
        A helper function to take the input data and proposed parameter sample and return the prediction
        Input:     data: (N x num_features) array of data
            theta: (w,v,b_h,b_o) vector of parameters with weights and biases
        '''
        self.decode(theta)  # method to decode w into W1, W2, B1, B2.
        size = x_data.shape[0]
        fx = np.zeros(size)
        for i in range(0, size):  # to see what fx is produced by your current weight update
            fx[i] = self.forward_pass(x_data[i,])
        return fx
        
    def sigmoid(self, x):
        # sigmoid activation function
        return 1 / (1 + np.exp(-x))  
        
    def softmax(self, x):
        #softmax function
        prob = np.exp(x) / np.sum(np.exp(x))
        return prob
\end{lstlisting}
\end{figure*}

\begin{figure*}[htbp]
\begin{lstlisting}[language=Python, label={lst:nn-forward-backward}, caption=Python implementation of Neural Network forward and backward passes]
# NN prediction
def forward_pass(self, X):
    '''
    Take an input X and return the output of the network
    Input:
        - X: (N x num_features) array of input data
    Output:
        - self.l2_output: (N) array of output data f(x) which can be 
        compared to observations (Y)
    '''
    # Hidden layer
    l1_z = np.dot(X, self.l1_weights) + self.l1_biases
    self.l1_output = self.sigmoid(l1_z) # activation function g(.)
    # Output layer
    l2_z = np.dot(self.l1_output, self.l2_weights) + self.l2_biases
    self.l2_output = self.sigmoid(l2_z)
    return self.l2_output

def backward_pass(self, X, Y):
    '''
    Compute the gradients using a backward pass and undertake Langevin-gradient updating of parameters
    Input:
        - X: (N x num_features) array of input data
        - Y: (N) array of target data
    '''
    # dE/dtheta
    l2_delta = (Y - self.l2_output) * (self.l2_output * (1 - self.l2_output))
    l2_weights_delta = np.outer(
        self.l1_output,
        l2_delta
    )
    # backprop of l2_delta and same as above
    l1_delta = np.dot(l2_delta,self.l2_weights.T) * (self.l1_output * (1 - self.l1_output))        
    l1_weights_delta = np.outer(
        X,
        l1_delta
    )

    # update for output layer
    self.l2_weights += self.lrate * l2_weights_delta
    self.l2_biases += self.lrate * l2_delta
    # update for hidden layer
    self.l1_weights += self.lrate * l1_weights_delta
    self.l1_biases += self.lrate * l1_delta
\end{lstlisting}
\end{figure*}

\begin{figure*}[htbp]

\begin{lstlisting}[language=Python, label={lst:langevin}, caption= Langevin-gradient functions in the Neural Network class]
    def langevin_gradient(self, x_data, y_data, theta, depth):
        '''
        Compute the Langevin-gradient proposal distribution 
        Input:
            - x_data: (N x num_features) array of input data
            - y_data: (N) array of target data
            - theta: (w,v,b_h,b_o) vector of proposed parameters.
            - depth: SGD depth
        Output: 
            - theta_updated: Updated parameter proposal
        '''
        self.decode(theta)  # method to decode w into W1, W2, B1, B2.
        size = x_data.shape[0] 
        # Update the parameters based on LG 
        for _ in range(0, depth):
            for ii in range(0, size):
                self.forward_pass(x_data[ii,])
                self.backward_pass(x_data[ii,], y_data[ii])
        theta_updated = self.encode()
        return  theta_updated
        
    def encode(self):
        '''
        Encode the model parameters into a vector 
        theta: vector of parameters.
        '''
        w1 = self.l1_weights.ravel()
        w2 = self.l2_weights.ravel()
        theta = np.concatenate([w1, w2, self.l1_biases, self.l2_biases])
        return theta

    def decode(self, theta):
        '''
        Decode the model parameters from a vector
        theta: vector of parameters.
        '''
        w_layer1size = self.input_num * self.hidden_num
        w_layer2size = self.hidden_num * self.output_num
        w_layer1 = theta[0:w_layer1size]
        self.l1_weights = np.reshape(w_layer1, (self.input_num, self.hidden_num))

        w_layer2 = theta[w_layer1size:w_layer1size + w_layer2size]
        self.l2_weights = np.reshape(w_layer2, (self.hidden_num, self.output_num))
        self.l1_biases = theta[w_layer1size + w_layer2size:w_layer1size + w_layer2size + self.hidden_num]
        self.l2_biases = theta[w_layer1size + w_layer2size + self.hidden_num:w_layer1size + w_layer2size + self.hidden_num + self.output_num]
\end{lstlisting}
\end{figure*}

\subsubsection{Python Implementation}

We first define and implement the simple neural network module (class), and implement methods (functions) for the  forward and backward pass to calculate the output of the network given a set of inputs. We need to compute the gradients and update the model parameters given a model prediction and observations, respectively. Listings \ref{lst:nn-model}, \ref{lst:nn-forward-backward}, and \ref{lst:nn-mcmc-sampler} presents the implementation \footnote{\href{https://github.com/sydney-machine-learning/Bayesianneuralnetworks-MCMC-tutorial/blob/main/04-Bayesian-Neural-Network.ipynb}{https://github.com/sydney-machine-learning/Bayesianneuralnetworks-MCMC-tutorial/blob/main/04-Bayesian-Neural-Network.ipynb}} of the Bayesian Neural Network and associated Langevin MCMC sampling scheme. Note that we implement the Bayesian Neural Network via the MCMC sampler class to sample (train) the weights and biases of the Neural Network class.

Next, we implement the model for a single hidden layer neural network with multiple input neurons and multiple output neurons (for binary and multi-class classification and multi-output regression). \textcolor{black}{Listing \ref{lst:nn-model} defines the Neural Network class with the \textit{constructor} function (\textit{init}) which defines the network topology, in terms of the number of input, hidden and output neurons along with the learning rate. These values are passed by the calling function. Next, we compute the total number of parameters by Line 17.  In Line 22, we initialize the network by calling the function (\textit{initialise\_network}) where we initialize (create) matrices for the weights from the input-hidden, and hidden-output layer, along with the vectors for their biases (Lines 33-49). Line 51 gives the \textit{evaluate\_proposal} function that takes the input data and proposed parameters (\textit{x\_data} and \textit{theta}) and \textit{returns} the prediction (\textit{fx}) in Line 62.  We feature the sigmoid transfer (activation) function in Line 64.}

\textcolor{black}{Listing \ref{lst:nn-forward-backward} lists the rest of the functions from the Neural Network class, where \textit{forward\_pass} propagates the information forward from input - hidden layer and then hidden to output layer  using a \textit{dot product} (Lines 12 and 16) and the \textit{returns} the output layer (Line 17). The \textit{backward\_pass} function begins by computing the gradients (delta) at the output layer (Line 27) and hidden layer (Line 33). Lines 40–44 update the weights in the hidden and output later using their respective gradients. }

\textcolor{black}{In a conventional backpropagation algorithm implementation, basically the forward and backward pass functions will be called in an iterative loop that will call these functions until the maximum number of epochs, or a given training (validation) error, has been reached. However, in our case, we are using the Langevin MCMC sampler to train the neural network model; hence, we have additional helper functions (Listing \ref{lst:langevin}) to ensure that the MCMC sampler class gets the information as needed. Essentially, from the MCMC class (Listing \ref{lst:nn-mcmc-sampler}), the sampler function (Listing \ref{lst:classmcmc}) calls the respective functions to evaluate the likelihood and the prior (Listing \ref{lst:bnn-lhood}). In the case of computing the likelihood,  the \textit{evaluate\_proposal} function in Listing \ref{lst:nn-model}  calls the decode function in Listing \ref{lst:langevin} to insert the values of the proposal (\textit{theta}) into the weight matrices and bias vectors of the model  defined in the \textit{NeuralNetwork} class of  Listing \ref{lst:nn-model}. }

\textcolor{black}{Next, we move to  Listing \ref{lst:nn-mcmc-sampler} that implements the Langevin MCMC sampler for classification problems, as given in the notebook of Github repository \footnote{\url{https://github.com/sydney-machine-learning/Bayesianneuralnetworks-MCMC-tutorial/blob/main/04a-Bayesian-Neural-Network-Classification.ipynb}}. We note that we also provide the implementation for regression/prediction problems in the notebook \footnote{\url{https://github.com/sydney-machine-learning/Bayesianneuralnetworks-MCMC-tutorial/blob/main/04-Bayesian-Neural-Network.ipynb}}. Furthermore, we also provide Python code implementation that features both classification and regression problems in the repository \footnote{\url{https://github.com/sydney-machine-learning/Bayesianneuralnetworks-MCMC-tutorial/tree/main/code}}. }

\textcolor{black}{In Listing \ref{lst:nn-mcmc-sampler}, we define the MCMC class with number of samples (\textit{n\_samples}), the burnin period  (\textit{n\_burnin}), along with the training (\textit{x\_data} and \textit{y\_data}) and test datasets (\textit{x\_test} and \textit{y\_test}). Lines 12-14 initializes the hyperparameters, such as the \textit{step\_size} of the random-walk on \textit{theta} and the \textit{sigma\_squared} that defined the spread of the Gaussian prior for the weights and biases. Lines 17-21 define the neural network model, the use of Langevin-gradients, the probability (\textit{l\_prob}) for using it, and the total number of weights and biases (\textit{theta\_size}). Next comes the storage of the parameters that are samples (Lines 24-25). Line 27 defines the function for \textit{model\_draws} - this is used post-sampling as a means to test the trained model. Line 50 defines the classification prediction accuracy, note that other error metrics can also be added.  }

\textcolor{black}{Listing \ref{lst:bnn-lhood} defines the multinomial Log-likelihood and prior for classification problems. Line 1 implements the multinomial log-likelihood function that uses the parameters and data (Equation \ref{multinomial}). Line 22 implements the log-prior function that uses the proposals (\textit{theta}) and the user-defined variance (\textit{sigma\_squared}) for the Gaussian prior. Note that in the case of regression and prediction problems, the log-likelihood and log-prior are similar to the Bayesian linear regression (Listing \ref{lst:gaussian-likelihood-prior}) with the omission of terms related to $\tau^2$.}


\begin{figure*}[htbp]
\begin{lstlisting}[language=Python, label={lst:nn-mcmc-sampler}, caption=Bayesian neural network using MCMC sampler for classification problems]  

class MCMC:
    def __init__(self, model, n_samples, n_burnin, x_data, y_data, x_test, y_test):
        self.n_samples = n_samples # number of MCMC samples
        self.n_burnin = n_burnin # number of burn-in samples
        self.x_data = x_data # (N x num_features)
        self.y_data = y_data # (N x 1)
        self.x_test = x_test # (Nt x num_features)
        self.y_test = y_test # (Nt x 1)

        # MCMC parameters - defines how much variation you need in changes to theta, tau
        self.step_theta = 0.025;   
        # Hyperpriors
        self.sigma_squared = 25 

        # initisalise the neural network model class
        self.model = model
        self.use_langevin_gradients = True
        self.sgd_depth = 1
        self.l_prob = 0.5 # likelihood prob
        self.theta_size = self.model.n_params # weights for each feature and a bias term

        # store output
        self.pos_theta = None
        self.rmse_data = None

    def model_draws(self, num_draws = 10, verbose=False):
        ''' Calculate the output of the network from draws of the posterior distribution
        Input: num_draws: number of draws,   verbose: if True, print the details of each draw
        Output: pred_y: (num_draws x N) ouptut of the NN for each draw'''
        accuracy = np.zeros(num_draws)
        pred_y = np.zeros((num_draws, self.x_data.shape[0]))
        sim_y = np.zeros((num_draws, self.x_data.shape[0]))

        for ii in range(num_draws):
            theta_drawn = np.random.normal(self.pos_theta.mean(axis=0), self.pos_theta.std(axis=0), self.theta_size)
            [likelihood_proposal, pred_y[ii,], sim_y[ii,], accuracy[ii]] = self.likelihood_function(
                theta_drawn
            )
            if verbose:
                print(
                    'Draw {} - accuracy: {:.3f}. Theta: {}'.format(
                        ii, accuracy[ii], theta_drawn
                    )
                )
        return pred_y, sim_y

    # Additional error metric
    @staticmethod
    def accuracy(predictions, targets):
        '''
        Additional error metric - accuracy
        '''
        count = (predictions == targets).sum()
        return 100 * (count / predictions.shape[0])
\end{lstlisting}
\end{figure*}

\begin{figure*}[htbp]
\begin{lstlisting}[language=Python, label={lst:bnn-lhood}, caption=Multinomial Log-likelihood and prior  for  classification  problems (Continued from Listing 11)]  
def likelihood_function(self, theta, test=False):
    ''' Calculate the multinomial log-likelihood of the data given the parameters and model
    Input:  theta: vector of parameters 
    Output:  log_likelihood: log likelihood of the data given the parameters '''
    if test:
        x_data = self.x_test
        y_data = self.y_test
    else:
        x_data = self.x_data
        y_data = self.y_data
    model_prediction, probs = self.model.evaluate_proposal(x_data, theta)
    model_simulation = model_prediction 
    accuracy = self.accuracy(model_prediction, y_data) #Accuracy error metric 
    # now calculate the log-likelihood
    log_likelihood = 0
    for ii in np.arange(x_data.shape[0]):
        for jj in np.arange(self.model.output_num):
            if y_data[ii] == jj:
                log_likelihood += np.log(probs[ii,jj])    
    return [log_likelihood, model_prediction, model_simulation, accuracy] 

def prior(self, sigma_squared, theta): 
    '''
    Calculate the prior of the parameters
    Input: sigma_squared - variance of normal prior for theta 
    Output: log_prior
    '''
    n_params = self.theta_size # number of parameters in model
    part1 = -1 * (n_params / 2) * np.log(sigma_squared)
    part2 = 1 / (2 * sigma_squared) * (sum(np.square(theta)))
    log_prior = part1 - part2
    return log_prior


\end{lstlisting}
\end{figure*}

\begin{figure*}[htbp]
\begin{lstlisting}[language=Python, label={lst:classmcmc}, caption=Implementation of MCMC sampler function  (Continued from Listing 12)]
 # MCMC sampler
def sampler(self):
    '''
    Run the sampler for a defined Neural Network model
    '''
    # define empty arrays to store the sampled posterior values
    # posterior of all weights and bias over all samples
    pos_theta = np.ones((self.n_samples, self.theta_size)) 

    # record output f(x) over all samples
    pred_y = np.zeros((self.n_samples, self.x_data.shape[0]))
    # record simulated values f(x) + error over all samples 
    sim_y = np.zeros((self.n_samples, self.x_data.shape[0]))
    # record the RMSE of each sample
    accuracy_data = np.zeros(self.n_samples)
    # now for test
    test_pred_y = np.ones((self.n_samples, self.x_test.shape[0]))
    test_sim_y = np.ones((self.n_samples, self.x_test.shape[0]))
    test_accuracy_data = np.zeros(self.n_samples)

    ## Initialisation
    # initialise theta
    theta = np.random.randn(self.theta_size)
    # make initial prediction
    pred_y[0,], _ = self.model.evaluate_proposal(self.x_data, theta)

    # Hyperparameters of priors - considered by looking at distribution of  similar trained  models - i.e distribution of weights and bias
    sigma_squared = self.sigma_squared

    # calculate the prior
    prior_val = self.prior(sigma_squared, theta)
    # calculate the likelihood considering observations
    [likelihood, pred_y[0,], sim_y[0,], accuracy_data[0]] = self.likelihood_function(theta)

    
\end{lstlisting}
\end{figure*}

\textcolor{black}{In Listing \ref{lst:loopbegins}, we begin sampling by first initializing variables that track the number of accepted proposals and how many times  Langevin-gradients are utilized,  this is just for analysis. In Line 4, we begin the sampling using a for loop that begins with 1 and ends with the number of samples. In Line 6, we propose the new values for the parameters (\textit{theta}) using random-walk proposal distribution centred at the mean of 0 and given spread (\textit{step\_size}) which needs to be experimentally determined in trial experiments. Then, we decide if we wish to use the Langevin-gradients or random-walk proposal distribution (Lines 7-8). Lines 9-20 implement the Langevin-gradients where we get one-step gradients from the neural network model. Therefore, we need to run a forward-pass and a backward-pass using the new sets of the current weights and biases, obtain the gradients of the output and hidden layers of the network, and concatenate to return these as a vector (Lines 15-20) of Listing \ref{lst:langevin}.}
  
\textcolor{black}{In Listing \ref{lst:loopbegins},  we then use the gradient (\textit{theta\_grad}) as the centre for the normal distribution to draw and add Gaussian noise to the gradients (Line 10). Then we again obtain the gradients, this is merely for the detailed balance condition. In Line 11, we get the gradients again, but this time we use the new values of theta (\textit{theta\_proposal}) that we earlier obtained in Line 6.  As given in Equation \ref{met_rat}, in the case when the proposals are not symmetric (i.e., Langevin-gradients), we need to get the q-ratio (Line 19  \textit{diff\_prop}). In the log-scale, this is obtained by the difference in the current \textit{theta} (first) and the new \textit{theta\_proposal} (second) to account for the detailed balance condition as shown in Line 19. In order to obtain the q-ratio, we need to further evaluate the old and the new proposals using the multivariate normal distribution and for numerical stability. However,  we need to have a simplified implementation for the multivariate distribution given a large set of weights and biases (\textit{theta}). Since we are operating in the log-scale, we can further simplify the multivariate normal distribution as shown in Lines 14-19.  Finally, Line  23 implements the case when random-walk proposal distribution would be used, note that in Line 22, the \textit{diff\_prop} is 0. This accounts for the detailed balance condition, since the proposals are naturally symmetric, in the case of random-walk proposal distribution (Line 23).}

\textcolor{black}{Next, we compute the log-likelihood in Listing \ref{lst:loopbegins} (Line 27) and the error metrics for the test dataset (Line 29). We note that this is a classification problem, and hence the classification accuracy  (Listing \ref{lst:bnn-lhood} - Line 13) is reported using the log-likelihood function. We determine the Metropolis-Hastings (MH) acceptance rate using Lines 34-35.  In Lines 32 and 33, we get the difference (ratio) for the proposed likelihood and the current likelihood, and the ratio for the prior with the current and proposed value of the prior. We utilize these to get the MH probability in Line 34, which also utilizes the difference (ratio) of proposed and current proposals, obtained either from Line 19 or Line 22. Finally, we either accept (Lines 39-45) or reject (Line 48) the proposal by comparing the MH probability with a random value obtained in Line 35. In the case if the proposal is accepted, the proposed values of theta along with prior and likelihood, becomes the current value in the chain. In the case if it is rejected, then the chain maintains its last accepted value as the current value (Line 48).  We remove the burn-in portion and store the posterior (Lines 53-57). Finally, we return the dictionary of the data that features the posterior and predictions using the Pandas library.}

\begin{figure*}[htbp]
\begin{lstlisting}[language=Python, label={lst:loopbegins}, caption= Begin with sampling loop (Continued from Listing 13)]   
    n_accept = 0  
    n_langevin = 0
    # Run the MCMC sample for n_samples
    for ii in tqdm(np.arange(1,self.n_samples)):
        # Sample new values for theta
        theta_proposal = theta + np.random.normal(0, self.step_theta, self.theta_size)
        lx = np.random.uniform(0,1,1)
        if (self.use_langevin_gradients is True) and (lx < self.l_prob):  
            theta_gd = self.model.langevin_gradient(self.x_data, self.y_data, theta.copy(), self.sgd_depth)  
            theta_proposal = np.random.normal(theta_gd, self.step_theta, self.theta_size)
            theta_proposal_gd = self.model.langevin_gradient(self.x_data, self.y_data, theta_proposal.copy(), self.sgd_depth) 
            # for numerical reasons, we will provide a simplified implementation that simplifies
            # the MVN of the proposal distribution
            wc_delta = (theta - theta_proposal_gd) 
            wp_delta = (theta_proposal - theta_gd)
            sigma_sq = self.step_theta
            first = -0.5 * np.sum(wc_delta * wc_delta) / sigma_sq 
            second = -0.5 * np.sum(wp_delta * wp_delta) / sigma_sq
            diff_prop =  first - second
            n_langevin += 1
        else:
            diff_prop = 0
            theta_proposal = np.random.normal(theta, self.step_theta, self.theta_size)
        # calculate the prior
        prior_proposal = self.prior(sigma_squared, theta_proposal)  # takes care of the gradients
        # calculate the likelihood considering observations
        [likelihood_proposal, pred_y[ii,], sim_y[ii,], accuracy_data[ii]] = self.likelihood_function(theta_proposal)
        # calculate the test likelihood
        [_, test_pred_y[ii,], test_sim_y[ii,], test_accuracy_data[ii]] = self.likelihood_function(
            theta_proposal, test=True)
        # since we using log scale: based on https://www.rapidtables.com/math/algebra/Logarithm.html
        diff_likelihood = likelihood_proposal - likelihood
        diff_prior = prior_proposal - prior_val
        mh_prob = min(1, np.exp(diff_likelihood + diff_prior + diff_prop))
        u = np.random.uniform(0, 1)
        # Accept/reject
        if u < mh_prob:
            # Update position
            n_accept += 1
            # update
            likelihood = likelihood_proposal
            prior_val = prior_proposal
            theta = theta_proposal
            # and store
            pos_theta[ii,] = theta_proposal
        else:
            # store
            pos_theta[ii,] = pos_theta[ii-1,]
    # print the % of times the proposal was accepted
    accept_ratio = (n_accept / self.n_samples) * 100
    print('{:.3}% were acepted'.format(accept_ratio))
    # store the posterior of theta
    self.pos_theta = pos_theta[self.n_burnin:, ]
    # Create a pandas dataframe to store the posterior samples of theta
    results_dict = {'w{}'.format(_): self.pos_theta[:, _].squeeze() for _ in range(self.theta_size-2)}
    results_dict['b0'] = self.pos_theta[:, self.theta_size-2].squeeze()
    results_dict['b1'] = self.pos_theta[:, self.theta_size-1].squeeze()    
    results_df = pd.DataFrame.from_dict(results_dict)
    return results_df
\end{lstlisting}
\end{figure*}

\subsection{Results}

We use the Sunspot time series\footnote{\url{https://www.sidc.be/silso/datafiles}} data and Abalone\footnote{\url{https://archive.ics.uci.edu/ml/datasets/abalone}} datasets for regression problems. The Abalone dataset provides the ring age for Abalone based on eight features that represent physical properties such as length, width, and weight and associated target feature, i.e., the ring age.  We note that determining the age of Abalone is difficult, as  requires cutting the shell  and counting the number of rings using a microscope. However, other physical  measurements can be used to predict the age and  a model can be developed to use the physical features to determine the ring age. Sunspots are regions of reduced surface temperature in the Sun's photosphere caused by concentrations of magnetic field flux, and  appears as spots darker than the surrounding areas. The  Sunspot  cycles are about every eleven years and over the solar cycle, the number of   Sunspot changes more rapidly. Sunspot activities are monitored since they have an include on Earth's climate and weather systems. We obtain the Abalone dataset from the University of California (UCI) Machine Learning Repository\footnote{\url{https://archive-beta.ics.uci.edu/about}} and keep a processed version of all the datasets in our repository\footnote{\href{https://github.com/sydney-machine-learning/Bayesianneuralnetworks-MCMC-tutorial/tree/main/data}{https://github.com/sydney-machine-learning/Bayesianneuralnetworks-MCMC-tutorial/tree/main/data}}. 

In the Sunspot time series problem, we employ a  one-step ahead prediction and hence use one output neuron. We process the Sunspot dataset (univariate time series) using Taken's embedding theorem \cite{Takens1981} to construct a state-space vector, also known as \textit{data windowing}. This is essentially using a sliding window approach of size $D$ overlapping $T$ time lags. The window size $D$ determines the number of input neurons in the Bayesian neural network and Bayesian linear model. We used $D=4$ and $T=2$ for our data  reconstruction for the Sunspot time series, as these values have given good performance in our previous works \cite{Chandra2019NC}. 

We also obtain datasets for classification problems from the same repository that features a large number of datasets for classification problems.    We selected the Iris classification dataset that contains 4 features (sepal length, sepal width, petal length, petal with) of three types of Iris flower species, featuring 50 instances for each case \footnote{\url{https://archive.ics.uci.edu/ml/datasets/iris}}. This dataset is one of the most prominent datasets used for machine learning.  We also selected the Ionosphere dataset that features a binary classification task with 351 instances \footnote{\url{https://archive.ics.uci.edu/ml/datasets/ionosphere}}. It has 34 continuous features and the task is to filter the radio signals as "good" or "bad".

In the Bayesian linear model and neural network, we choose the number of samples to be 25,000 for all problems, distributed across 5 chains and excluding 50\% burn-in. In the Bayesian linear model, we choose the learning rate  $r=0.1$, and the step sizes for $\theta=0.02$ and $\tau=0.01$, respectively. Additionally, for the Gaussian prior distribution, we choose the parameters $\sigma^2 =5, \nu_1=0$ and $\nu_2=0$, respectively. In the Bayesian neural network models, we choose the learning rate $r=0.01$, and the step sizes for $\theta=0.025$ and $\tau=0.2$, respectively. In the Gaussian prior distribution, we choose the hyperparameter $\sigma^2=25$ determined from examining trained neural network models for similar problems. In the case of regression, we use inverse-Gamma prior for $\tau^2$, and hence use hyperparameters for this prior, $\nu_1=0$ and $\nu_2=0$, respectively. We also use a burn-in rate of 0.5 for both the Bayesian linear and the Bayesian neural network models.   In the case of Bayesian neural networks, we apply Langevin-gradients at a rate of 0.5. 

\begin{figure*}[htbp!]
\centering
\begin{subfigure}{0.6\textwidth}
\centering
\resizebox{1\linewidth}{!}{\input{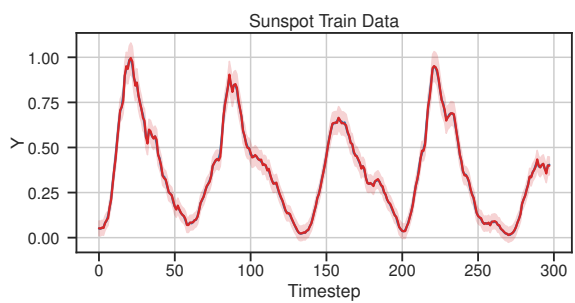}}
\caption{Predictions and actual train dataset for Sunspot}
\end{subfigure}
\begin{subfigure}{0.3\textwidth}
\centering
\resizebox{1\linewidth}{!}{\input{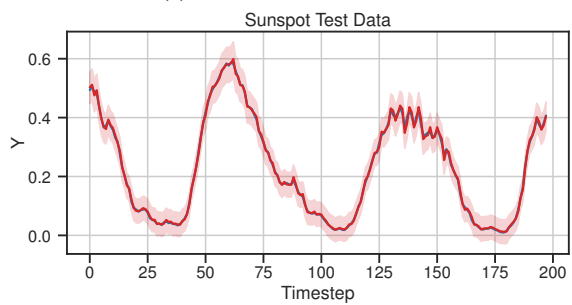}}
\caption{Predictions and actual of $\Delta Y$ train dataset for Sunspot prediction.}
\end{subfigure}
\vspace{2em}
\begin{subfigure}{0.6\textwidth}
\centering
\resizebox{1\linewidth}{!}{\input{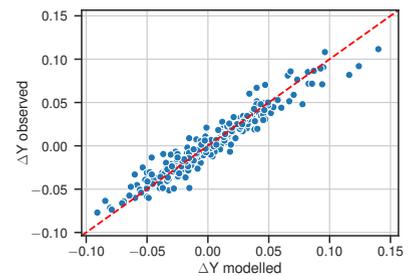}}
\caption{Predictions and actual test dataset for Sunspot}
\end{subfigure}
\begin{subfigure}{0.3\textwidth}
\centering
\resizebox{1\linewidth}{!}{\input{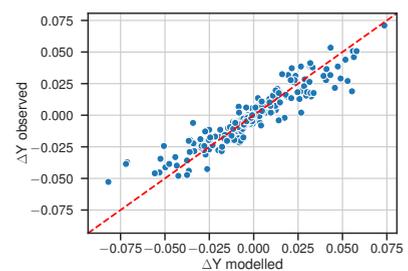}}
\caption{Predictions and actual of $\Delta Y$ test dataset for Sunspot prediction.}
\end{subfigure}
\caption{Bayesian linear regression model - Sunspot dataset}
\label{sunspot_result_plot_blr}
\end{figure*}
\begin{figure*}[htbp!]
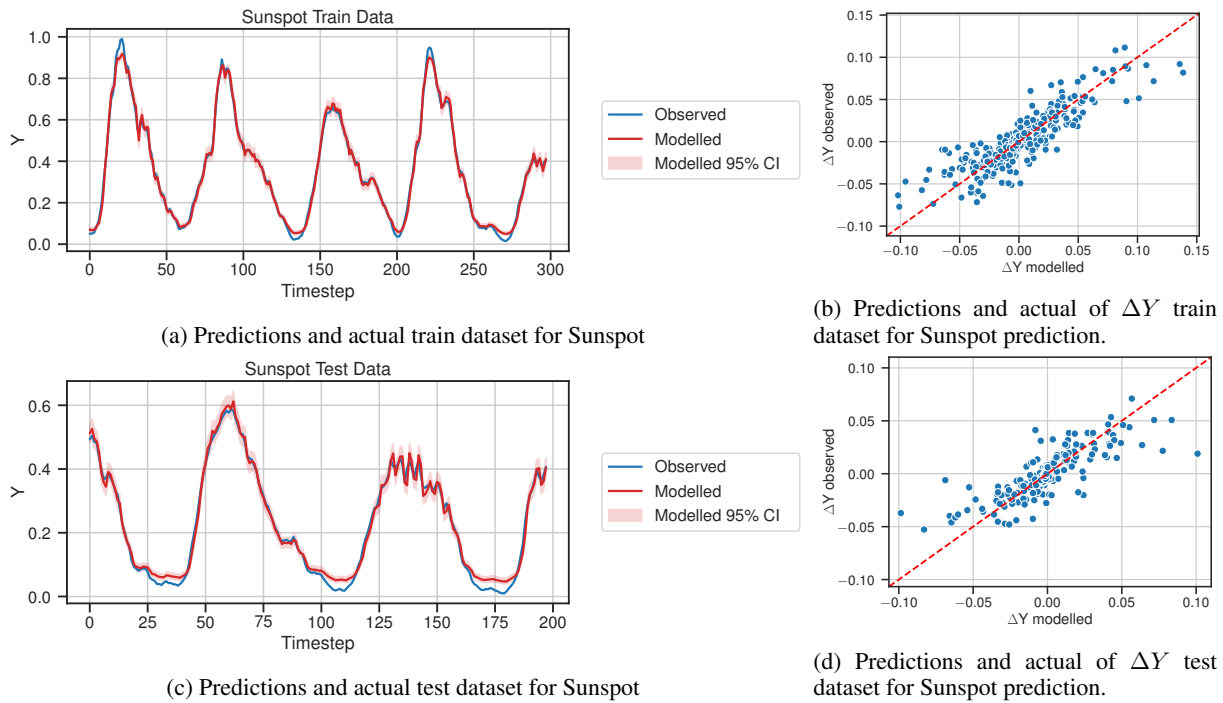

\centering
\begin{subfigure}{0.6\textwidth}
\centering
\resizebox{1\linewidth}{!}{\input{figures/results/bnn/Sunspot_2025_train.pgf}}
\caption{Predictions and actual train dataset for Sunspot}
\end{subfigure}
\begin{subfigure}{0.3\textwidth}
\centering
\resizebox{1\linewidth}{!}{\input{figures/results/bnn/Sunspot_2025_train_dy.pgf}}
\caption{Predictions and actual of $\Delta Y$ train dataset for Sunspot prediction.}
\end{subfigure}
\vspace{2em}
\begin{subfigure}{0.6\textwidth}
\centering
\resizebox{1\linewidth}{!}{\input{figures/results/bnn/Sunspot_2025_test.pgf}}
\caption{Predictions and actual test dataset for Sunspot}
\end{subfigure}
\begin{subfigure}{0.3\textwidth}
\centering
\resizebox{1\linewidth}{!}{\input{figures/results/bnn/Sunspot_2025_test_dy.pgf}}
\caption{Predictions and actual of $\Delta Y$ test dataset for Sunspot prediction.}
\end{subfigure}
\caption{Bayesian neural network model - Sunspot dataset}
\label{sunspot_result_plot_bnn}
\end{figure*}

We first present the results of Bayesian regression with the Sunspot (time series prediction) and Abalone (regression) datasets. We evaluate the model performance using the \textit{root mean squared error} (RMSE) which is a standard metric for time series prediction and regression problems. \textcolor{black}{We present the results obtained by the Bayesian linear model and Bayesian neural network model for regression problems in Table \ref{regression_result_table}. Figures \ref{sunspot_result_plot_blr} and  \ref{sunspot_result_plot_bnn} present the prediction plots (observed, modelled and 95 \% credible interval)  that show a comparison between Bayesian linear model and neural network models for the fixed training (Panels a and b) and test (Panels c and d) datasets. We note that Panels a and c show the timeseries prediction (x-axis represents timestep number). Panels b and d present a scatter-plot of the change from timestep $t-1$ to $t$ in the observed ($\Delta Y$ observed) and predicted values ($\Delta Y$  modelled). This gives an indication of model's ability to predict change at each timestep with a skill better than persistence (as observed $Y_{t-1}$ is given as an input to the model, a model predicting $y_t = Y_{t-1}$ could have a low RMSE). Thus, we note that the RMSE does not assess the skill (capability)  of the model above the observed $Y_{t-1}$ (given these models are conducting one step-ahead prediction); however, given the analysis using Panels b and d, the performance of the two models can be further compared using RMSE.} 

In Table \ref{regression_result_table}, we observe that the Bayesian neural network performs better for the Sunspot time series prediction problem, as it achieves a better accuracy (lower RMSE) on both the training and testing set. This can also be seen in Figures \ref{sunspot_result_plot_blr} and \ref{sunspot_result_plot_bnn}. In the case of the Abalone problem, Table \ref{regression_result_table} shows that both models obtain similar classification performance, but  the Bayesian neural network has better test performance. However, we note that in both problems, Bayesian neural networks have a much lower acceptance rate; we prefer roughly a 23 \% acceptance rate  \cite{gelman1997weak}  that implies that the posterior distribution has been effectively sampled.  However, we also note that such MCMC sampling acceptance rates are typically based on statistical and linear models, which may not apply to Bayesian neural networks. Therefore,  more research needs to be done to determine a good acceptance rate that aligns with convergence and \textit{ergodicity} \cite{Chris2006}.

\begin{table*}[htbp!]
    \centering
    \renewcommand{\arraystretch}{1.2}
    \begin{tabular}{c c c c c} 
        \hline
    Method  &Problem & Train RMSE & Test RMSE & Accept. rate \\ 
        \hline
    \multirow{2}{*}{Bayesian linear model} & \multirow{2}{*}{Sunspot} & 0.025 &      0.022  & 13.5\%  \\ 
    & & (0.013) & (0.012) & \\
    \multirow{2}{*}{Bayesian neural network} & \multirow{2}{*}{Sunspot} &  0.027 &      0.026  & 7.4\% \\ 
    & & (0.007) & (0.007) & \\
    \hline
   \multirow{2}{*}{Bayesian linear model} & \multirow{2}{*}{Abalone} & 0.085 &      0.086 & 5.8\% \\ 
   & & (0.005) & (0.005) & \\
   \multirow{2}{*}{Bayesian neural network} & \multirow{2}{*}{Abalone} & 0.080 &      0.080  & 3.8\%   \\ 
    & & (0.002) & (0.002) & \\
        \hline
    \end{tabular}

    \caption{Results using Bayesian linear model and Bayesian neural networks via MCMC sampler for the Abalone (regression)  and the Sunspot (prediction) problems. The results show the RMSE mean and standard deviation (in brackets)  for the train and test datasets, respectively.} 
    \label{regression_result_table}
\end{table*}

Table \ref{classification_result_table} presents results for the classification problems in the Iris and Ionosphere datasets. We notice that both models have similar test and training classification performance for the Iris classification problem, and Bayesian neural networks give better results for the test dataset for the Ionosphere problem. The acceptance rate is much higher for Bayesian neural networks in the case of classification, and in the case of regression, only the Bayesian linear model has a more suitable acceptance rate.
   


        

\begin{table*}[htbp!]
    \centering
    \renewcommand{\arraystretch}{1.2}
    \begin{tabular}{c c c c c} 
        \hline
    Method  &Problem & Train  Accuracy  & Test  Accuracy  & Accept. rate \\ 
        \hline
    \multirow{2}{*}{Bayesian linear model} & \multirow{2}{*}{Iris} & 90.392\% &     90.844\% & 83.5\%  \\ 
    & & (2.832) & (3.039) & \\
    \multirow{2}{*}{Bayesian neural network} & \multirow{2}{*}{Iris} &  97.377\% &     98.116\% & 97.0\% \\ 
    & & (0.655) & (1.657) & \\
    \hline
   \multirow{2}{*}{Bayesian linear model} & \multirow{2}{*}{Ionosphere} & 89.060\% &     85.316\% & 58.8\% \\ 
   & & (1.335) & (2.390) & \\
   \multirow{2}{*}{Bayesian neural network} & \multirow{2}{*}{Ionosphere} & 99.632\% &     92.668\% & 94.5\%   \\ 
    & & (0.356) & (1.890) & \\
        \hline
    \end{tabular}
    \caption{Classification accuracy with Bayesian linear model and Bayesian neural networks via MCMC. The results show the accuracy mean and standard deviation (in brackets)  for  the train and test datasets.}
    \label{classification_result_table}
\end{table*}

\subsection{Convergence diagnosis}

It is important to ensure that the MCMC sampling is adequately exploring the parameter space and constructing an accurate picture of the posterior distribution. One method of monitoring the performance of the adopted MCMC sampler is to examine \textit{convergence diagnostics} that assess/monitor the extent to which the Markov chains have become a stationary distribution. Practitioners routinely apply the Gelman-Rubin (GR) convergence diagnostic  \cite{gelman1992inference} that is developed by sampling from multiple MCMC chains, whereby the variance of each chain is assessed independently (within-chain variance) and then compared to the variance between the multiple chains (between-chain variance) for each parameter. A large difference between these two variances would indicate that the chains have not converged on the same stationary distribution.

In our case, we run several independent experiments and compare the MCMC chains using a modified Gelman-Rubin convergence diagnostic presented by Vehtari et al. \cite{vehtariRankNormalizationFoldingLocalization2021}. It is beyond the scope of this publication to provide a detailed mathematical description of the convergence diagnostics, the reader can refer to \cite{vehtariRankNormalizationFoldingLocalization2021} for a full description. A useful package for Bayesian model diagnostics, which contains an implementation of this modified diagnostic is \textit{arviz} \cite{kumarArviZUnifiedLibrary2019} \footnote{\url{https://python.arviz.org/en/stable/api/generated/arviz.rhat.html}}. We refer to the modified Gelman-Rubin diagnostic as $\hat{R}$, where values close to 1 indicate convergence. In Listing \ref{lst:gelman-rubin}, we present the code to prepare the MCMC sampler outputs for convergence diagnostic by \textit{arviz}.




\begin{figure*}[htbp]
\begin{lstlisting}[language=Python, label={lst:gelman-rubin}, caption=Convergence diagnostics using the \textit{arviz} python package that could be run after Listing 7]
    import arviz as az
    # Sample a second chain as more than one chain is required to generate the Rhat diagnostic
    # Run second chain using a different seed
    np.random.seed(2)
    # setup second chain
    mcmc_chain2 = MCMC(n_samples, burn_in, x_data, y_data, x_test, y_test)
    # Run the sampler
    results_chain2 = mcmc_chain2.sampler()

    # Now combine these two chains by stacking and convert to arviz dataset
    # To convert to arviz, convert the pandas dataframe of samples into a dictionary of lists, stack and ingest into arviz
    res_dict = results.to_dict(orient='list')
    res_dict_chain2 = results_chain2.to_dict(orient='list')
    # stack the chains looping through each parameter and ingest into arviz
    az_results = az.from_dict(
        {par: np.vstack([res_dict[par], res_dict_chain2[par]]) for par in res_dict}
    )

    # now we can use arviz to obtain summary statistics with Rhat ("r_hat") for each parameter.
    az.summary(az_results)
\end{lstlisting}
\end{figure*}

\subsubsection{Results}

\begin{figure}[htbp!]
    \centering
    \resizebox{0.9\linewidth}{!}{\input{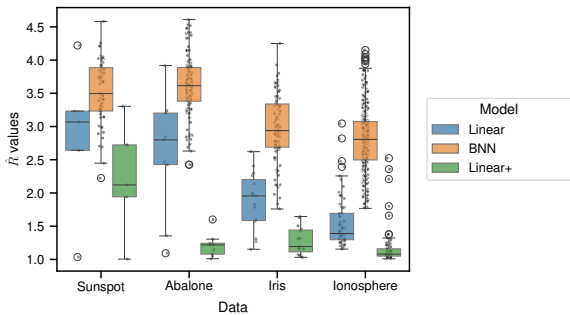}}
    \caption{Distribution of $\hat{R}$ values for our Bayesian linear model (Linear), Bayesian neural network (BNN) and a linear model with 10 times more samples (Linear+). The scattered points show the underlying data summarised in the box plots. The "Linear+" case shows the ability of the sampler to converge as more samples are taken.}
    \label{r_hat_distribution}
    \end{figure}

We show results for the modified Gelman-Rubin diagnostic for Bayesian linear and Bayesian neural network models for each of the four datasets.
We test five chains for each model, each with 5,000 samples of weights (excluding 50\% burn-in samples) and computing  the $\hat{R}$ values for each parameter. We also provide an additional example "Linear+" for the linear models where each of the 5 chains has 50,000 samples (excluding 50\% burn-in samples). The basic sampler presented here is not state of the art in terms of sampling efficiency (see Section \ref{sec:discussion} for further discussion), and for some of the problems (particularly those where parameters are difficult to identify) it may take a large number of samples to converge. This additional example is shown to demonstrate that the convergence is improving as the number of samples grows. Figure \ref{r_hat_distribution} shows the distribution of the $\hat{R}$ values, and we observe that the $\hat{R}$ values of the weights for the Bayesian linear regression model are much smaller than Bayesian neural network.  We can observe  that based on the Gelman-Rubin diagnostics, the Bayesian neural network shows poor convergence. The additional samples in the "Linear+" case improve convergence, in particular for the Abalone, Iris and Ionosphere cases.

\begin{figure*}[htbp!]
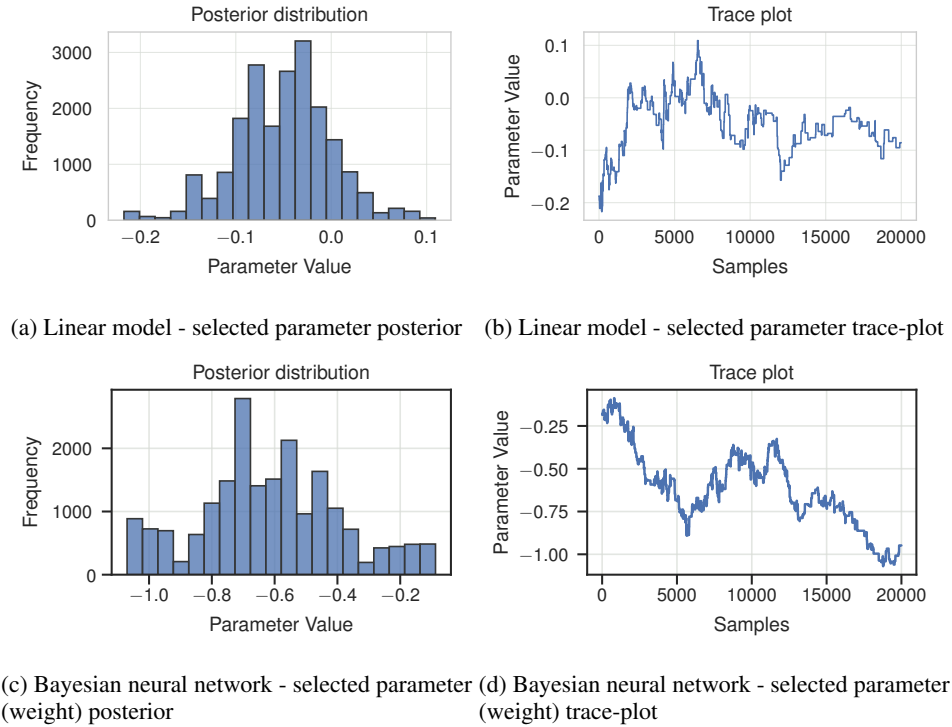

\centering
\begin{subfigure}{0.35\linewidth}
    \resizebox{\linewidth}{!}{\input{figures/results/linear_tp_nothinning_Sunspot_w0_posterior.pgf}}
    \caption{Linear model - selected parameter   posterior}
\end{subfigure}
\begin{subfigure}{0.35\linewidth}
    \resizebox{\linewidth}{!}{\input{figures/results/linear_tp_nothinning_Sunspot_w0_trace.pgf}}
    \caption{Linear model - selected parameter   trace-plot}
\end{subfigure}
\begin{subfigure}{0.35\linewidth}
    \resizebox{\linewidth}{!}{\input{figures/results/bnn_tp_nothinning_Sunspot_w1_posterior.pgf}}
    \caption{Bayesian neural network - selected parameter (weight)  posterior}
\end{subfigure}
\begin{subfigure}{0.35\linewidth}
    \resizebox{\linewidth}{!}{\input{figures/results/bnn_tp_nothinning_Sunspot_w1_trace.pgf}}
    \caption{Bayesian neural network - selected parameter (weight)  trace-plot}
\end{subfigure}

\caption{Posterior and trace-plot for a parameter in each of the Bayesian linear and Bayesian neural network models - Sunspot data. In this example, 20,000 samples were taken excluding 20\% burn-in with no thinning.}
\label{post_trace_plot_convergence_lm}
\end{figure*}

\begin{figure*}[htbp!]
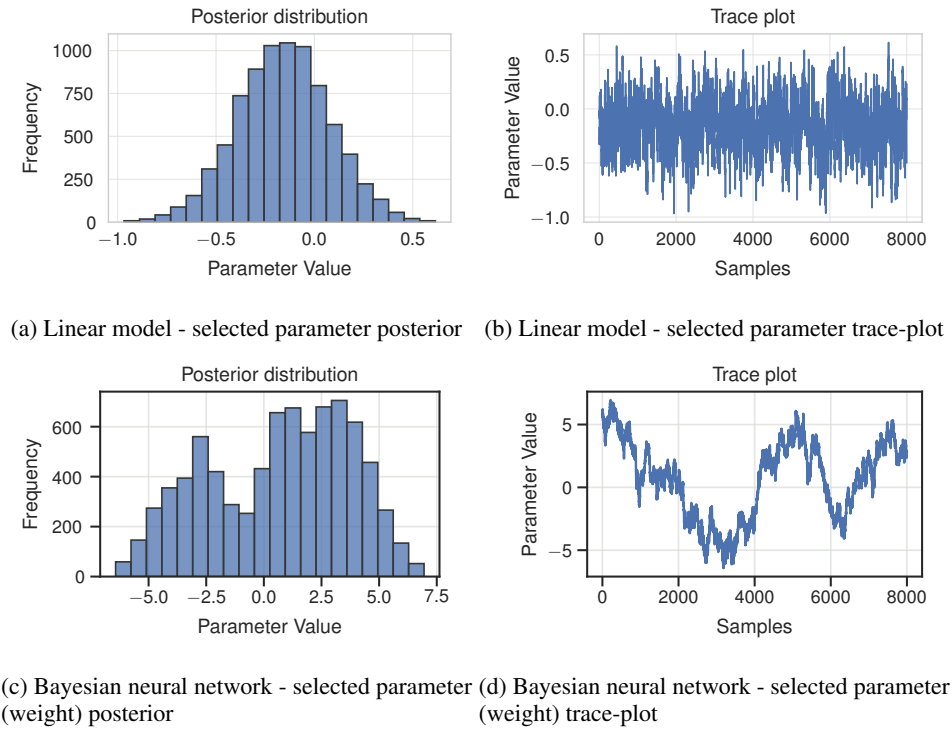

\centering
\begin{subfigure}{0.35\linewidth}
    \resizebox{\linewidth}{!}{\input{figures/results/linear_tp_Iris_w5_posterior.pgf}}
    \caption{Linear model - selected parameter posterior}
\end{subfigure}
\begin{subfigure}{0.35\linewidth}
    \resizebox{\linewidth}{!}{\input{figures/results/linear_tp_Iris_w5_trace.pgf}}
    \caption{Linear model - selected parameter   trace-plot}
\end{subfigure}
\begin{subfigure}{0.35\linewidth}
    \resizebox{\linewidth}{!}{\input{figures/results/bnn_tp_Iris_b1_posterior.pgf}}
    \caption{Bayesian neural network - selected parameter (weight)  posterior}
\end{subfigure}
\begin{subfigure}{0.35\linewidth}
    \resizebox{\linewidth}{!}{\input{figures/results/bnn_tp_Iris_b1_trace.pgf}}
    \caption{Bayesian neural network - selected parameter (weight)  trace-plot}
\end{subfigure}

\caption{Posterior and trace-plot for a selected parameter in each of the Bayesian linear and Bayesian neural network models - Iris data. We took 400,000 samples  (excluding 20\% burn-in) with the chain thinned by a factor of 50 for visualisation.}
\label{post_trace_thinning_convergence}
\end{figure*}

By closely examining each weight individually, we observe that the problem of non-convergence mainly arises from the multi-modality of the posterior. In Figure \ref{post_trace_plot_convergence_lm}, we look at samples from a single chain of 20,000 samples excluding 20\% burn-in. In Figure \ref{post_trace_plot_convergence_lm}a and \ref{post_trace_plot_convergence_lm}b, we present a visualization for a selected weight from the Bayesian linear model. In Figure \ref{post_trace_plot_convergence_lm}c and \ref{post_trace_plot_convergence_lm}d, we present a visualization for a selected weight from the Bayesian neural network model. We observe potential multi-modal distributions in both cases, with a high degree of auto-correlation and poor convergence. To examine the impact of longer MCMC chains in achieving convergence, we ran an additional test by taking 400,000 samples excluding 20\% burn-in, and then thinning the chain by a factor of 50 for visualisation. \textcolor{black}{Thinning is the process of reducing the memory burden of the chain, particularly where samples may be auto-correlated \cite{linkThinningChainsMCMC2012}. We can also use thinning to more easily visualise long chains with many samples. In our case, we implement thinning by retaining every 50th sample of the chain.} We present these results for the Iris dataset in Figure \ref{post_trace_thinning_convergence}. We can see that the chains exhibit more desirable properties,  particularly in the case of the linear model (Figure \ref{post_trace_thinning_convergence}- Panels a and b). We can see in Figure \ref{post_trace_thinning_convergence} - Panels c and d, that the Bayesian neural network exhibits a multimodel posterior for this parameter. 

\textcolor{black}{Furthermore, other approaches for convergence diagnosis such as autocorrection analysis can also be implemented with packages such as the \textit{integrated autocorrelation time} \cite{wolff2004monte} in \textit{Emcee} \cite{foreman2013emcee} \footnote{\url{https://emcee.readthedocs.io/en/stable/tutorials/autocorr/}}. We refer to readers for a comprehensive review of MCMC convergence diagnostics given by \cite{cowles1996markov,roy2020convergence,south2022postprocessing}.}



 \subsection{MCMC packages}

 \textcolor{black}{We note that there are avenues to further improve the sampling efficiency, we utilized Langevin MCMC but other gradient-based approaches such as Hamiltonian MCMC (HMC)\cite{neal2011mcmc} also exist. It is worthwhile to evaluate the performance of Langevin MCMC against other advanced gradient-based sampling algorithms (e.g., No-U-turn Sampler (NUTS) \cite{hoffman2014no}) to assess convergence properties in cases of multimodal posterior distributions, such as in Bayesian neural networks. We note that  implementation of HMC and NUTS exist in probabilistic programming libraries such as PyMC \cite{patil2010pymc} and Stan \cite{carpenter2017stan,annis2017bayesian}.} We implemented an additional notebook using the NumPyro \cite{phanComposableEffectsFlexible2019} probabilistic programming library to perform an equivalent Bayesian linear regression (Listing 3-7) as an example \footnote{\url{https://github.com/sydney-machine-learning/Bayesianneuralnetworks-MCMC-tutorial/blob/main/05-Linear-Model_NumPyro.ipynb}}.

\subsection{Discussion}
\label{sec:discussion}

We presented a Python tutorial for Bayesian neural networks and Bayesian linear models using MCMC sampling. 
In general, we observed that the Bayesian neural network performs better than  Bayesian linear regression in our selected problems (Tables 1 and 2), despite showing no poor convergence (Figure 10, Figures and  12-Panel d). This could be due to the challenge of sampling a relatively large number of weights and biases of Bayesian neural networks, which also have multimodal posterior distribution. Hence, we conclude that for the case of Bayesian neural networks, a poor performance in the Gelman-Rubin diagnostics does not necessarily imply a poor performance in prediction tasks.  We revisit the principle of \textit{equifinality}\cite{cicchetti1996equifinality,gresov1997equifinality} which states that in open systems, a given end state can be reached by many potential means. In our case, the system is a neural network model and many solutions exist that represent a trained model displaying an accepted level of performance accuracy. However, we note that Bayesian models offer uncertainty quantification in predictions, and proper convergence is required. The original Gelman-Rubin diagnosis \cite{gelman1992inference} motivated several enhancements for different types of problems \cite{brooks1998general,roy2020convergence,vats2021revisiting, vehtariRankNormalizationFoldingLocalization2021}; and we may need to develop a better diagnosis for Bayesian neural networks. Nonetheless, in our case comparing Bayesian logistic regression (converged) and Bayesian neural networks (not converged but achieved good accuracy), we can safely state that the Bayesian neural network presented can only provide a means for uncertainty quantification, but is not mature enough to qualify as a robust Bayesian model.  

We also revisit the convergence issue in the case of the Bayesian linear model as shown in Panel (b) - Figure \ref{post_trace_plot_convergence_lm}, which shows a multi-modal distribution that has not well converged.   It may also be the case that certain features are not contributing much to the decision-making process (predictions) and the weights associated (coefficients) with those features may be difficult to sample. This is similar to the case of neural networks, where certain weight links are not needed and can be pruned. Pruning neural networks create compact models that also  get better generalization performance \cite{liang2021pruning}. 

We note that we attained a much higher acceptance rate in Bayesian neural networks for classification problems  (Table \ref{classification_result_table}) when compared to regression problems(Table \ref{classification_result_table}). We note that different likelihood functions are used for classification and regression (Multinomial and Gaussian likelihood) and we utilized the Langevin-gradient proposal distributions that accounted for the detailed balanced condition. Hence, we need to further fine-tune the hyperparameters associated with the proposal distribution to ensure we get a higher acceptance rate for the regression problems. Fine-turning the hyperparameters for the proposal distribution is a laborious task which can be seen as a major limitation of MCMC sampling in large models such as Bayesian neural networks. Although 23 \% acceptance rate \cite{gelman1997weak}  has been prominently used as a "golden rule", the optimal acceptance rate depends on the nature of the problem. The number of parameters, Langevin-based proposal distribution, and type of model would raise questions about the established acceptance rate \cite{bedard2008optimal}. Hence, more work needs to be done to establish what acceptance rates are appropriate for simple neural networks and deep learning models. 

A way to address the issue of convergence would be to develop an ensemble of linear models that can compete with the accuracy of neural networks or deep learning models. In ensemble methods such as bagging and boosting, we can use linear models that have attained convergence as per Gelman-Rubin diagnosis and then combine the results of the ensemble using averaging and voting, as done in ensemble methods.   

Our previous work has shown that despite the challenges, the combination of Langevin-gradients with parallel tempering MCMC \cite{Chandra2019NC}, presents opportunities for sampling larger neural network architectures such as autoencoders and graph-based CNNs\cite{chandra2021bayesian,chandra2021revisiting}. The need to feature a robust methodology for uncertainty quantification in CNNs will make them more suitable for applications where uncertainty in decision-making poses major risks, such as medical image analysis \cite{anwar2018medical} and human security \cite{akcay2018security}. CNNs have been considered for modelling temporal sequences, they have proven to be successful for time series classification \cite{zhao2017convolutional,liu2018time}, and time series forecasting problems \cite{borovykh2017conditional,binkowski2018autoregressive,chandra2021evaluation}. It has also been shown that one-dimensional CNNs provide better prediction performance than the conventional LSTM network for multi-step ahead time series prediction problems \cite{chandra2021evaluation}. Leveraging CNNs within a Bayesian framework can provide better uncertainty quantification in predictions and make them useful for cutting-edge real-world applications. We need a comprehensive evaluation  of prominent gradient-based MCMC sampling  methods for deep learning models such as CNNs, autoencoders,  and LSTM networks.

 We envision that this tutorial will enable statisticians, machine learning and deep learning experts to utilize MCMC sampling more effectively when developing new models and  Bayesian frameworks for existing deep learning models. The tutorial has introduced basic concepts with code and provides an overview of challenges when it comes to the convergence of Bayesian neural networks. It can be further extended to utilize parallel computing via tempered MCMC \cite{Chandra2019NC}, HMC for Bayesian neural networks, and Langevin MCMC and HMC for Bayesian deep learning.



\subsection*{Code and Data}

All code (for implementation, results and figures) and data presented in this paper are available in the associated github repository\footnote{\href{https://github.com/sydney-machine-learning/Bayesianneuralnetworks-MCMC-tutorial}{https://github.com/sydney-machine-learning/Bayesianneuralnetworks-MCMC-tutorial}}. This repository presents the implementations in separate Jupyter notebooks in the base directory, with sub-directories containing data, convenient functions and details of the environment setup.

\subsection*{Acknowledgements and contributions}

R. Chandra contributed by conceptualisation, coding and experiments, analysis, writing, and project supervision.  J. Simmons contributed by coding, analysis, and writing. 

J. Simmons is funded by the Australian Research Council ITTC in Data Analytics for Resources and Environments (Grant IC19010031). We thank Royce Chen from UNSW Sydney for the initial experiments and analysis.

\bibliographystyle{ieeetr} 
\bibliography{2017,2018,Chandra-Rohitash,Bays,sample,cnnbayes,mcmc_tutorial}

 \begin{IEEEbiography}[{\includegraphics[width=1in,height=1.25in,clip,keepaspectratio]{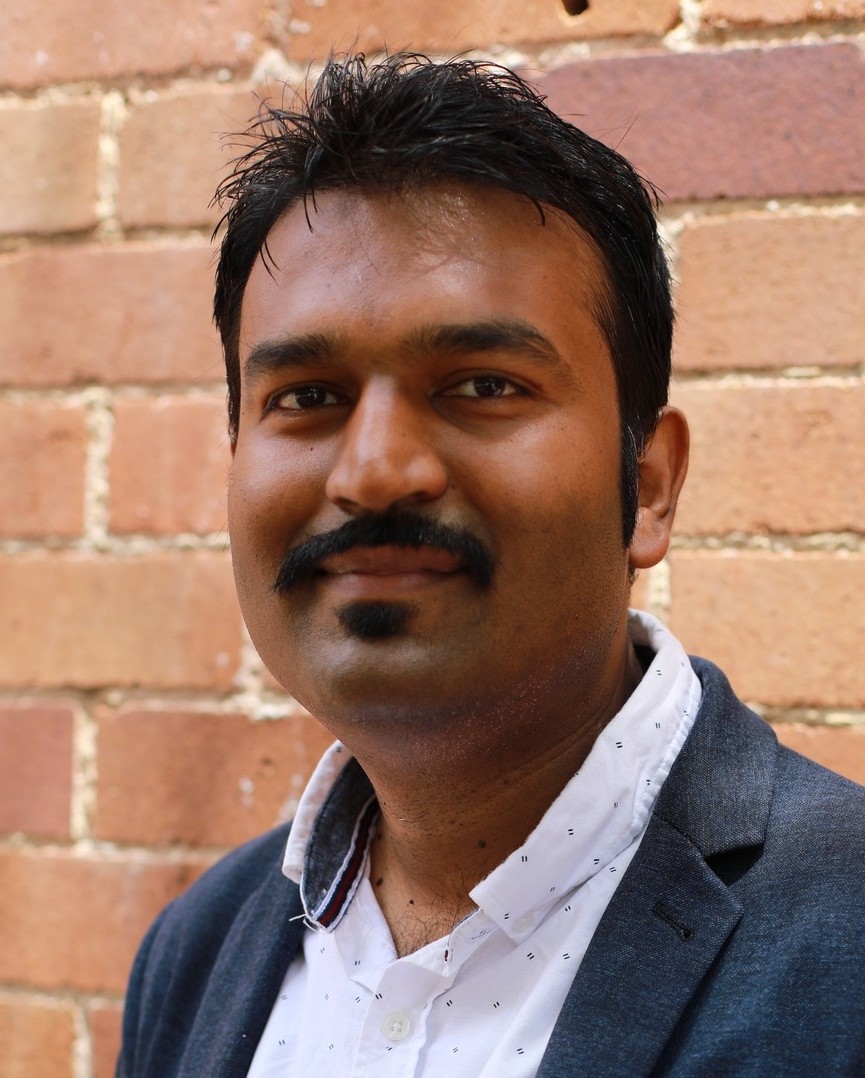}}] 
{Dr Rohitash Chandra} is a Senior Lecturer in Data Science at the UNSW School of Mathematics and Statistics.  He leads a program of research encircling methodologies and applications of artificial intelligence. The methodologies include  Bayesian deep learning, neuroevolution,  ensemble learning, and data augmentation. The applications include climate extremes, geoscientific models, mineral exploration, biomedicine, and language modelling. Dr Chandra is on the Editorial Board for Geoscientific Model Development. He also served as an Associate Editor for IEEE Transactions on Neural Networks and Learning Systems, and Neurocomputing (2021-2022).  Since 2021, Dr Chandra has been featured in the Stanford's list of top 2\% scientists in the world.
\end{IEEEbiography}


 \begin{IEEEbiography}[{\includegraphics[width=1in,height=1.25in,clip,keepaspectratio]{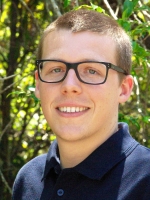}}] 
{Dr Joshua Simmons} received his Ph.D from UNSW Sydney and is currently working as a Research Engineer at the ARC Training Centre in Data Analytics for Resources and Environments (DARE). His research has included numerical and statistical modelling of coastal processes, with a particular focus on storm erosion of sandy beaches. More recently, his work has focused on interpretable machine learning/data science approaches and Bayesian methods for uncertainty quantification in the hydrology domain.
\end{IEEEbiography}

\EOD

\end{document}